\documentclass[nohyperref]{article}

\usepackage{microtype}
\usepackage[accepted]{icml2022}

\usepackage{mathtools}
\icmltitlerunning{Towards Demystifying Representation Learning with Non-contrastive Self-supervision}

\usepackage{tabularx}
\usepackage{graphicx}
\usepackage{hyperref}\usepackage{natbib}
\usepackage{delarray}
\usepackage{times}
\usepackage{subfigure}
\usepackage{color}
\usepackage{amssymb}
\usepackage{amsmath}
\usepackage{amsthm}
\usepackage{mathrsfs}
\usepackage{tikz}
\usepackage{xspace}
\usepackage{enumerate}
\usepackage{epstopdf}

\usepackage{algorithm}
\usepackage{algorithmic}
\usepackage{nameref}
\usepackage{booktabs}
\usepackage{mdframed}
\usepackage{fmtcount}
\usepackage{footnote}
\usepackage{bbm}
\usepackage{thmtools,thm-restate}
\newcommand{\poly}{{\text{poly}}}
\newcommand{\eat}[1]{}
\newcommand{\R}{{\mathbb R}}
\newcommand{\N}{{\mathcal N}}

\newcommand{\E}{{\mathbb{E}}}

\newcommand{\inner}[2]{\left\langle #1, #2 \right\rangle}
\newcommand{\ns}[1]{\left\| #1 \right\|^2}
\newcommand{\n}[1]{\left\| #1 \right\|}
\newcommand{\fns}[1]{\left\| #1 \right\|^2_F}
\newcommand{\fn}[1]{\left\| #1 \right\|_F}

\newcommand{\pr}[1]{\left( #1 \right)}
\newcommand{\br}[1]{\left[ #1 \right]}

\newcommand{\absr}[1]{\left| #1 \right|}

\newcommand{\whp}{\text{with probability at least  } 1-\exp(-\Omega(n))}
\newcommand{\tX}{\widetilde{X}}
\newcommand{\calN}{\mathcal{N}}
\newcommand{\calS}{S}

\newcommand{\hP}{\hat{P}}
\newcommand{\hw}{\hat{w}}
\newcommand{\tw}{\widetilde{W}}

\newcommand{\wat}{W_{a,t}}
\newcommand{\wpt}{W_{p,t}}
\newcommand{\wt}{W_t}
\newcommand{\twat}{\widetilde{W}_{a,t}}
\newcommand{\twpt}{\widetilde{W}_{p,t}}
\newcommand{\twt}{\widetilde{W}_t}

\newcommand{\stopgrad}{\text{StopGrad}}

\newcommand{\stepsize}{\gamma}

\renewcommand{\algorithmicrequire}{ \textbf{Input:}} %Use Input in the format of Algorithm
\renewcommand{\algorithmicensure}{ \textbf{Output:}} %Use Output in the format of Algorithm

        {\hspace*{\fill}$\Box$\par\vspace{4mm}}
\newenvironment{proofof}[1]{\smallskip\noindent{\bf Proof of #1.}}%
        {\hspace*{\fill}$\Box$\par}

\newenvironment{itemize*}%
{\begin{itemize}[leftmargin=*,topsep=0pt]%
		\setlength{\itemsep}{0pt}%
		\setlength{\parskip}{0pt}}%
	{\end{itemize}}
\newenvironment{enumerate*}%
{\begin{enumerate}[leftmargin=*,topsep=0pt]%
		\setlength{\itemsep}{0pt}%
		\setlength{\parskip}{0pt}}%
	{\end{enumerate}}

\newcommand\fb[1]{\textcolor{black}{#1}}

\newtheorem{theorem}{Theorem}
\newtheorem{lemma}[theorem]{Lemma}

\newtheorem{assumption}{Assumption}

\hypersetup{
    colorlinks=true,       % false: boxed links; true: colored links
    linkcolor=black,    %cyan,        % color of internal links
    citecolor=black,        % color of links to bibliography
    filecolor=magenta,     % color of file links
    urlcolor=blue
}

\usepackage{hyperref}
\usepackage{url}

\begin{document}

\twocolumn[
\icmltitle{Towards Demystifying Representation Learning with Non-contrastive Self-supervision}



\icmlsetsymbol{equal}{*}

\begin{icmlauthorlist}
\icmlauthor{Xiang Wang}{duke}
\icmlauthor{Xinlei Chen}{fair}
\icmlauthor{Simon S. Du}{fair,uw}
\icmlauthor{Yuandong Tian}{fair}
\end{icmlauthorlist}

\icmlaffiliation{duke}{Duke University}
\icmlaffiliation{uw}{University of Washington}
\icmlaffiliation{fair}{Facebook AI Research}

\icmlcorrespondingauthor{Xiang Wang}{xwang@cs.duke.edu}
%\icmlcorrespondingauthor{Firstname2 Lastname2}{first2.last2@www.uk}

\icmlkeywords{Machine Learning, ICML}

\vskip 0.3in
]



\printAffiliationsAndNotice{}  % leave blank if no need to mention equal contribution

\def\ncssl{nc-SSL\xspace}
\def\ours{DirectCopy\xspace}
\def\directset{\text{DirectSet}(\alpha)\xspace}

\begin{abstract}
Non-contrastive methods of self-supervised learning (such as BYOL and SimSiam) learn representations by minimizing the distance between two views of the same image. 
These approaches have achieved remarkable performance in practice, but the theoretical understanding lags behind.
\citet{tian2021understanding} explained why the representation does not collapse to zero, however, how the feature is learned still remains mysterious. In our work, we prove in a linear network, non-contrastive methods learn a desirable projection matrix and also reduce the sample complexity on downstream tasks. Our analysis suggests that weight decay acts as an implicit threshold that discards the features with high variance under data augmentations, and keeps the features with low variance. Inspired by our theory, we design a simpler and more computationally efficient algorithm DirectCopy by removing the eigen-decomposition step in the original DirectPred algorithm in \citet{tian2021understanding}. Our experiments show that DirectCopy rivals or even outperforms DirectPred on STL-10, CIFAR-10, CIFAR-100 and ImageNet.
\end{abstract}

\section{Introduction}

Self-supervised learning emerges as a promising direction to learn representations without manual labels. As one popular approach, contrastive learning~\citep{oord2018representation,tian2019contrastive,bachman2019learning,he2020momentum,simclr} minimizes the distances between representations of two augmented views of the same data point (positive pairs), and maximizes such distances between the views of different data points (negative pairs). Intuitively, minimizing distances between positive pairs encourages the learned representation to be invariant under data augmentations, and maximizing distances between negative pairs helps avoid representational collapse (i.e., mapping all the data to the same representation). 

Recently, \emph{non-contrastive} self-supervised learning (abbreviated as \textbf{\emph{\ncssl}}) was proposed to learn representations using only positive pairs. Presumably, \ncssl might converge to the trivial constant representation that is a global minimizer of the loss function. However, in practice, \ncssl is able to learn nontrivial representation and shows remarkable performance on downstream tasks (e.g., image classification~\citep{byol,chen2020exploring}). This brings about two fundamental questions: (1) without negative pairs, why the learned representation does not collapse to trivial (i.e., constant) solutions, and (2) what  representation \ncssl learns from the training and how the learned representation reduces the sample complexity in downstream tasks.  

\begin{figure}[t]
\centering
\includegraphics[width=0.9\linewidth]{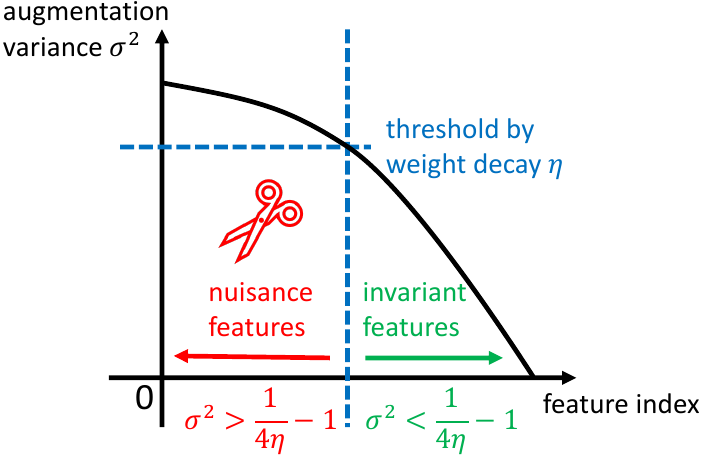}
\caption{\small In non-contrastive self-supervised learning (abbr. \textbf{\emph{\ncssl}}), the weight decay acts as an implicit threshold on the variance of features under data augmentations. With weight decay, the training process discards nuisance features that have high variance under data augmentations and keeps invariant features that have low variance.}\label{fig:summary}
\vskip -0.15in
\end{figure}

While many theoretical results on contrastive SSL~\citep{arora2018a,lee2020predicting,tosh2020contrastive,wen2021toward} exist, similar study on \ncssl has been very rare. As one of the first work towards this direction, \citet{tian2021understanding} showed that while the global optimum of the non-contrastive loss is indeed a trivial one, following gradient direction in \ncssl{}, one can find a \emph{local} optimum that admits a nontrivial representation. Based on their theoretical findings on gradient-based methods, they proposed a new approach, DirectPred, that directly sets the predictor using the eigen-decomposition of the correlation matrix of inputs before the predictor, rather than updating it with gradient methods. As a method for \ncssl{}, DirectPred shows comparable or better performance in multiple datasets, including CIFAR-10~\citep{krizhevsky2009learning}, STL-10~\citep{coates2011analysis} and ImageNet~\citep{imagenet_cvpr09}, compared to BYOL~\citep{byol} and SimSiam~\citep{chen2020exploring} that optimize the predictor using gradient descent. 

While~\citet{tian2021understanding} addressed the first question, i.e., why the learned representation does not collapse to zero, they did not address the second question, i.e., how the training dynamics in \ncssl leads to a meaningful representation that depends on the data augmentations and reduces the sample complexity on down-stream tasks. 

\paragraph{Main Contributions.}
In this paper, we make a first attempt towards the second question, by studying a family of algorithms named $\mathbf{DirectSet}(\alpha)$, in which the DirectPred algorithm proposed by~\citet{tian2021understanding} is a special case with $\alpha=1/2$. Our contribution is two-fold:

First, we perform a theoretical analysis on $\directset$ with linear networks. We prove that $\directset$ learns a desirable projection matrix onto the invariant features given polynomial number of unlabeled samples. 
Our analysis shows that there exists an \emph{implicit threshold}, determined by weight decay parameter $\eta$, that governs which features are learned and which are discarded. As illustrated in Figure~\ref{fig:summary}, the threshold is applied to the variance of the feature across different data augmentations (or ``views'') of the same instance: \emph{nuisance features} (features with high variances under augmentation) are discarded, while \emph{invariant features} (i.e., with low variances) are kept. We further prove the learned representation can reduce the sample complexity on downstream tasks. To the best of knowledge, this is the first result proving \ncssl learns meaningful representations that reduce the sample complexity on downstream tasks.

Second, we show that $\mathbf{DirectCopy}$, a special case of $\directset$ when $\alpha=1$, performs comparably with (or even outperforms) DirectPred on downstream tasks in CIFAR-10, CIFAR-100, STL-10 and ImageNet. In \ours, the predictor can be set \emph{without} the expensive eigen-decomposition operation, which makes \ours much simpler and more efficient than DirectPred. %Besides, interesting behaviors are found in \ncssl beyond the linear model, e.g., the discarded features gradually come back after training over longer epochs. Such observations shed light on the feature learning process in deep nonlinear networks. We form a hypothesis with supporting evidences and leave a thorough study for future work. 

\paragraph{Organization.} In Section~\ref{sec:related_works}, we discuss the related works. We introduce DirectSet($\alpha$) and DirectCopy in Section~\ref{sec:prelim} and analyze them in a linear network setting in Section~\ref{sec:two_layer}.  Section~\ref{sec:empirical_perf} demonstrates the empirical performance of DirectCopy across various datasets and Section~\ref{sec:ablation} shows ablation experiments. Finally, we conclude the paper in Section~\ref{sec:conclusion}. 

\section{Related Works}\label{sec:related_works}

\paragraph{Contrastive methods:} Contrastive learning~\citep{oord2018representation,tian2019contrastive,bachman2019learning,he2020momentum,simclr} learns representations by minimizing the distances of positive pairs and maximizing distances of negative pairs. There are many theoretical works trying to explain contrastive learning~\citep{arora2018a,wang2020understanding,tian2020makes,tsai2020self,tosh2021contrastive}. \citet{haochen2021provable} proposed a contrastive loss that implicitly performs spectral decomposition on the augmentation graph. \citet{tian2020understanding} showed that gradient updates tend to amplify the features invariant to augmentations. \citet{wen2021toward} proved that data augmentations decouple the correlations between spurious dense features and force the network to learn desired sparse features. 

%\citet{arora2018a} assumed that different positive samples are drawn from the same latent class, which is essentially supervised learning. \citet{wang2020understanding} proved that the contrastive loss optimizes the alignment and uniformity of the features. \citet{tian2020makes} and \citet{tsai2020self} analyzed contrastive learning from an information-theoretical perspective. \citet{tosh2021contrastive} examined contrastive learning from the perspective of multi-view redundancy. \citet{haochen2021provable} proposed a contrastive loss that implicitly performs spectral decomposition on the augmentation graph. \citet{tian2020understanding} showed that gradient updates tend to amplify the features invariant to data augmentations. \citet{wen2021toward} proved that data augmentations decouple the correlations between spurious dense features and force the network to learn desired sparse features. 

\paragraph{Non-contrastive methods:} Without negative samples, non-contrastive methods use other techniques to avoid representational collapse. BYOL~\citep{byol} and SimSiam~\citep{chen2020exploring} use an extra predictor and a stop-gradient operation. SwAV~\citep{caron2020unsupervised} clusters the data while ensuring different views of the same data falls in the same cluster. \citet{zbontar2021barlow,bardes2021vicreg,hua2021feature} de-correlate the variables in the features. \citet{ermolov2021whitening} proposed a new loss function based on the whitening of the latent space features.  DINO~\citep{caron2021emerging} applies centering and sharpening on the target network outputs. In this work, we study BYOL and SimSiam as representative \ncssl methods.

\paragraph{Comparison with \citet{tian2021understanding}} \citet{tian2021understanding} only explained why the representation in nc-SSL does not collapse to zero, but did not study what representation is learned and how the representation is related to the data distribution and augmentation process. In particular, they assumed the augmentation is isotropic in all dimensions and did not define the invariant features and nuisance features. In our model, we relax the isotropic assumption and allow the augmentation to act only in the nuisance subspace. Our analysis for the first time explains the representation learning mechanism in nc-SSL: weight decay discards the nuisance features and keeps the invariant features.
Motivated by the analysis, we also design a simpler and more efficient algorithm (DirectCopy), which achieves comparable or even better performances than the original DirectPred proposed by~\citet{tian2021understanding}.

%In \ncssl, different techniques are proposed to avoid collapsing.  Beyond these, BatchNorm (including its variants~\citep{byol-without-bn}), de-correlation~\citep{zbontar2021barlow,bardes2021vicreg,hua2021feature}, whitening~\citep{ermolov2021whitening}, centering~\citep{caron2021emerging}, and online clustering~\citep{caron2020unsupervised} are all effective ways to enforce implicit contrastive constraints among samples for collapsing prevention. We study BYOL and SimSiam as representative \ncssl methods.

\section{Preliminaries}\label{sec:prelim}

%In this section, we first introduce useful notations and then present our algorithm: DirectSet($\alpha$) and DirectCopy.

\subsection{Notations}

We use $I_d$ to denote the $d\times d$ identity matrix and simply write $I$ when the dimension is clear. For any linear subspace $S$ in $\R^d$, we use $P_S \in \R^{d\times d}$ to denote the projection matrix on $S.$ More precisely, the projection matrix $P_S$ equals $UU^\top,$ where the columns of $U$ constitute a set of orthonormal bases for subspace $S$. We use $\N(\mu, \Sigma)$ to denote the Gaussian distribution with mean $\mu$ and covariance $\Sigma.$

We use $\n{\cdot}$ to denote spectral norm for a matrix, or $\ell_2$ norm for a vector and use $\fn{\cdot}$ to denote Frobenius norm for a matrix. 
For a real symmetric matrix $A\in \R^{d\times d}$ whose eigen-decomposition is $\sum_{i=1}^d \lambda_i u_i u_i^\top,$ we use $|A|$ to denote $\sum_{i=1}^d \absr{\lambda_i} u_i u_i^\top$. If $A$ is also positive semi-definite, we use $A^{\alpha}$ to denote $\sum_{i=1}^d \lambda_i^\alpha u_i u_i^\top$ for any positive $\alpha\in\R.$

\subsection{DirectSet($\alpha$) and DirectCopy}\label{sec:algorithm}

\begin{figure}[h]
\centering
\includegraphics[width=0.9\linewidth]{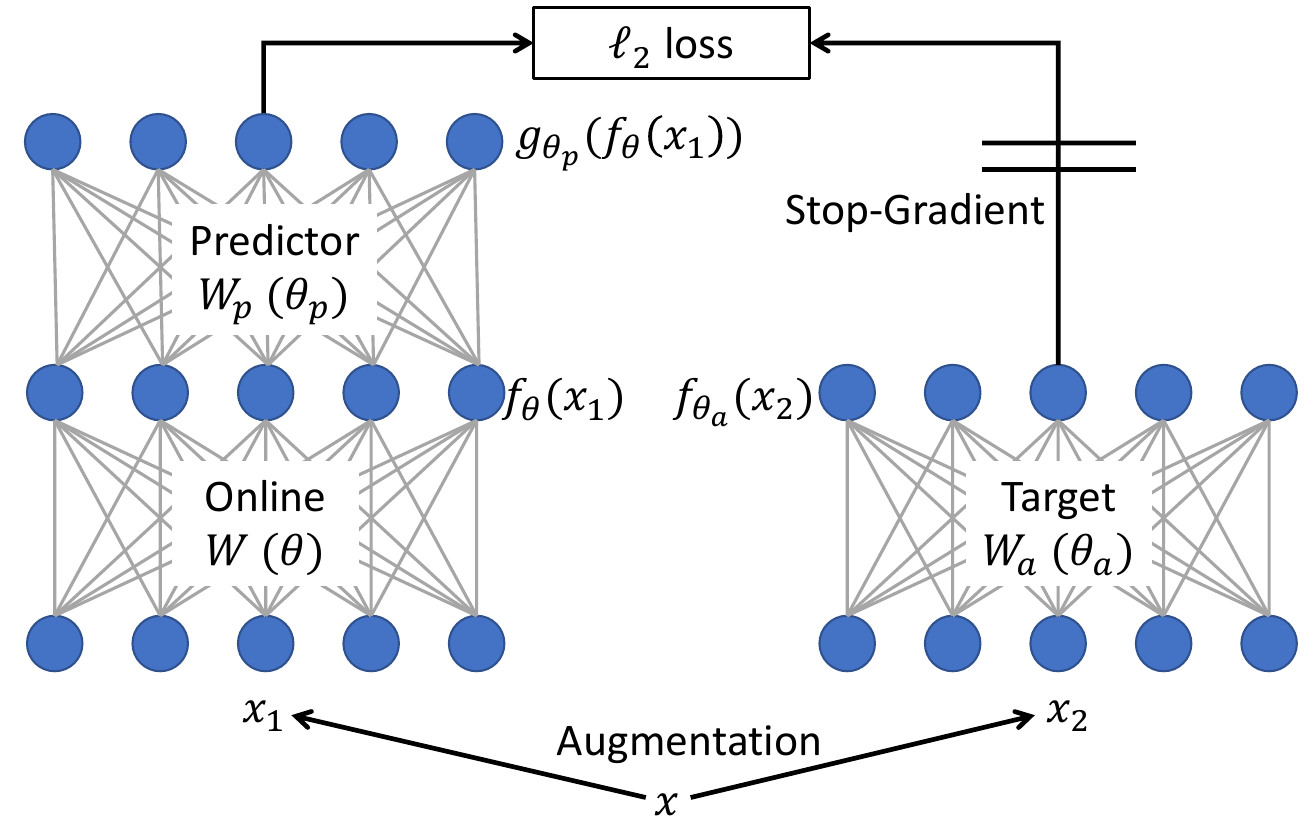}
\caption{\small Problem Setup of a linear network. Both BYOL and DirectSet($\alpha$) update online $W$ by gradient methods and set target $W_a$ as EMA of $W$. BYOL also updates predictor $W_p$ by gradient methods, while DirectSet($\alpha$) sets $W_p$ based on the correlation matrix of predictor inputs ($\E_{x_1}f_\theta(x_1)f_\theta(x_1)^\top$).
}

\label{fig:two_layer}
\end{figure}

In \ncssl, recent methods as BYOL~\citep{byol} and SimSiam~\citep{chen2020exploring} employ a dual pair of Siamese networks~\citep{Bromley1994-mk}: one side is a composition of an online network (including a projector) and a predictor network, the other side is a target network (see Figure~\ref{fig:two_layer} for a simple example). The target network has the same architecture as the online network, but has potentially different weights. Given an input $x$, two augmented views $x_1,x_2$ are generated, and the network is trained to match the representation of $x_1$ (through the online network and the predictor network) and the representation of $x_2$ (through the target network). More precisely, suppose the online network and the target network are two mappings $f_{\theta},f_{\theta_a}: \R^d \mapsto \R^h$ and the predictor network is a mapping $g_{\theta_p}: \R^h\mapsto \R^h$, the network is trained to minimize the following loss $L(\theta,\theta_p,\theta_a)$:
$$\frac{1}{2}\E_{x_1, x_2}\ns{\frac{g_{\theta_p}\pr{f_\theta(x_1) }}{\n{g_{\theta_p}\pr{f_\theta(x_1) }}}-\stopgrad\pr{\frac{f_{\theta_a}(x_2) }{\n{f_{\theta_a}(x_2)}}} }.$$
In BYOL and SimSiam, the online network and the target network are trained by running gradient methods on $L$. The target network is not trained by gradient methods; instead, it is directly set with the weights in the online network (SimSiam) or an exponential moving average (EMA) of the online network (BYOL).

\citet{tian2021understanding} proposed DirectPred that directly sets the predictor based on the correlation matrix of the predictor inputs. DirectPred achieves comparable performance as BYOL and admits much cleaner theoretical analysis. Therefore, in this paper, we focus our theoretical analysis on DirectSet($\alpha$) (a family of algorithms that include DirectPred as a special case), although we expect some of the insights also apply to BYOL/SimSiam.

Given a positive scalar $\alpha$, DirectSet($\alpha$) sets the predictor based on the correlation matrix $F$ of the predictor inputs:
$$W_p = \frac{F^\alpha}{\n{F^\alpha}}+\epsilon I,$$
where $F=\E_{x_1}f_\theta(x_1)f_\theta(x_1)^\top.$ 
In practice, $F$ is estimated by a moving average over batches. That is,
$$\hat{F} = \mu \hat{F} + (1-\mu)\E_{B}[f_\theta(x_1)f_\theta(x_1)^\top],$$
where $\E_{B}$ is the expectation over one batch. The predictor regularization $\epsilon I$, when properly chosen, can improve the quality of the learned representations (see the experiments and analysis in Section~\ref{sec:ablation}).

In the original DirectPred algorithm, $\alpha$ is fixed at $1/2$. To compute $\hat{F}^{1/2}$, one needs to first compute the eigen-decomposition of $\hat{F}$, and then taking the square root of each eigenvalue. This step of eigen-decomposition can be expensive especially when the representation dimension $h$ is high. To avoid the eigen-decomposition step, we propose DirectCopy ($\alpha=1$), in which the predictor $W_p$ is a direct copy of the $\hat{F}$ (with normalization and regularization)\footnote{Computing the spectral norm of $\hat{F}$ is much faster than computing the eigen-decomposition of $\hat{F}$, because the former only needs the top eigen-vector of $\hat{F}$. Table~\ref{tbl:normalization} shows that the spectral normalization can also be removed or be replaced by Frobenius normalization without hurting the performance.}. As we shall see, DirectCopy enjoys both theoretical guarantees and strong empirical performance.

\newcommand{\acc}{\hat{\epsilon}}

\section{Theoretical Analysis of DirectSet($\alpha$)}\label{sec:two_layer}

We prove DirectSet($\alpha$) learns meaningful representations and reduces sample complexity of down-stream tasks when the online/target network is a linear network. For simplicity, we focus on the setting where the online network is a single-layer network in this section, although our analysis also extends to deep linear networks (see Appendix~\ref{sec:deep_linear}). Deep linear networks have been widely used as a tractable theoretical model for studying nonconvex loss landscapes \citep{kawaguchi2016deep,du2019width,laurent2018deep} and nonlinear learning dynamics ~\citep{saxe2013exact,Saxe2019-eg,Lampinen2018-sl,arora2018optimization} in supervised learning setting. \citet{tian2021understanding} also analyzed \ncssl{} on a linear network, but did not analyze their proposed approach DirectPred.

\subsection{Setup}\label{sec:two_layer_setup}

In this subsection, we define the network model, data distribution and simplify DirectSet($\alpha$) algorithm for our theoretical analysis. We consider the following network model (see Figure~\ref{fig:two_layer}),
\begin{assumption}[Linear network model]\label{assump:network}
The online, predictor and target network are all single-layer linear network without bias, with weight matrices denoted as $W, W_p, W_a \in \R^{d\times d}$ respectively.
\end{assumption}

\begin{figure}[t]
\centering
\includegraphics[width=0.8\linewidth]{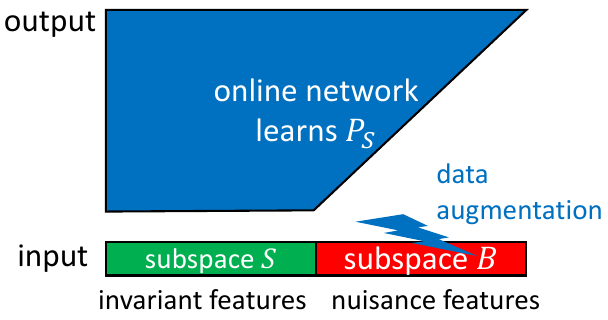}
\caption{\small The input space is a direct sum of subspace $S$ of invariant features and subspace $B$ of nuisance features. With the data augmentations only applying on subspace $B$, the online network converges to the projection matrix onto $S$ subspace after training.
}\label{fig:two_subspaces}
\end{figure}

For the data distribution, we assume the input space is a direct sum of an invariant feature subspace and a nuisance feature subspace (see Figure~\ref{fig:two_subspaces}).
Specifically, we assume
\begin{assumption}[Data distribution]\label{assump:data_distribution}
The input $x$ is sampled from $\N(0,I_d)$, and its augmented view $x_1, x_2$ are independently sampled from $\N(x, \sigma^2P_B),$ where $B$ is a $(d-r)$-dimensional subspace. We denote $S$ as the orthogonal subspace of $B$ in $\R^d.$
\end{assumption}
In this simple data distribution, subspace $S$ corresponds to the features that are invariant to augmentations and its orthogonal subspace $B$ is the nuisance subspace which the augmentation changes. We will prove that DirectSet($\alpha$) can learn the projection matrix onto $S$ subspace. Note in the previous work~\citep{tian2021understanding}, they assumed the covariance of the augmentation distribution to be $\sigma^2 I$ and did not study what representation is learned. 

\paragraph{Algorithm simplification: }For the convenience of analysis, we consider a simplified version of DirectSet($\alpha$). We compute the loss function without normalizing the two representations, so the population loss $L(W, W_a, W_p)$ is 
\begin{equation}
     \frac{1}{2}\E_{x_1,x_2} \ns{W_p W x_1 - \stopgrad\pr{W_a x_2}}\label{eqn:population_loss},
\end{equation}
and the empirical loss $\hat{L}(W,W_p,W_a)$ is
\begin{equation}
    \frac{1}{2n}\sum_{i=1}^n \ns{W_p Wx_1^{(i)} - \stopgrad(W_a x_2^{(i)})}\label{eqn:empirical_loss},
\end{equation}
where $x^{(i)}$'s are independently sampled from $\N(0,I)$, and augmented views $x_1^{(i)}$ and $x_2^{(i)}$ are independently sampled from $\N(x^{(i)}, \sigma^2 P_B)$. To train our model, we initialize the online network as a scaled identity matrix, which greatly facilitates our analysis. 
\begin{assumption}[Identity initialization]\label{assump:init}
The online network weight $W$ is initialized as $\delta I$ with $\delta$ a positive real number. 
\end{assumption}
We run gradient flow or gradient descent on online network $W$ with weight decay $\eta$, and set the the target network $W_a = W.$ For clarity of presentation, when training on the population loss, we set $W_p$ as $(W\E_x xx^\top W^\top)^\alpha = (WW^\top)^\alpha$ instead of $(W\E_{x_1} x_1x_1^\top W^\top)^\alpha$ as in practice; when training on the empirical loss, we set $W_p$ as $(W \frac{1}{n}\sum_{i=1}^n x^{(i)}[x^{(i)}]^\top W^\top)^\alpha.$ Here, we set the predictor regularization $\epsilon = 0$ and its influence will be studied in Section~\ref{sec:ablation}.

%In the following, DirectSet($\alpha$) is shown to recover the projection matrix $P_S$ with polynomial number of samples. Furthermore, given that the learned matrix is close to $P_S,$ the sample complexity on downstream tasks is reduced. 

\subsection{Gradient Flow on Population Loss}\label{sec:GF_population}
In this subsection, we show that DirectSet($\alpha$) running on the population loss with infinitesimal learning rate can learn the projection matrix onto the invariant feature subspace $S$.

\begin{restatable}{theorem}{learnsubspace}\label{thm:learn_subspace}
Suppose network architecture and data distribution follow Assumption~\ref{assump:network} and Assumption~\ref{assump:data_distribution}, respectively. Suppose we initialize online network $W$ as $\delta I,$ and run \fb{DirectSet$(\alpha)$} on population loss (see Eqn.~\ref{eqn:population_loss}) with infinitesimal step size and $\eta$
weight decay. If $\eta \in \pr{ \frac{1}{4(1+\sigma^2)}, \frac{1}{4}}$ and $\delta>\pr{\frac{1-\sqrt{1-4\eta}}{2}}^{1/(2\alpha)},$ then $W$ converges to $\pr{\frac{1+\sqrt{1-4\eta}}{2}}^{1/(2\alpha)}P_S$ when time goes to infinity.
\end{restatable}

Theorem~\ref{thm:learn_subspace} shows that when the weight decay is in certain range, and when the initialization is large enough, the online network can converge to the desired projection matrix $P_S$~\footnote{Note that Theorem~\ref{thm:learn_subspace} also holds with negative initialization $\delta<-\pr{\frac{1-\sqrt{1-4\eta}}{2}}^{1/(2\alpha)},$ in which case $W$ converges to $-\pr{\frac{1+\sqrt{1-4\eta}}{2}}^{1/(2\alpha)}P_S.$ Our other results can be extended to negative $\delta$ in a similar way.}. 
In sequel, we explain how the dynamics of $W$ leads to a projection matrix and how the weight decay and initialization scale come into play. We leave the full proof in Appendix~\ref{sec:proofs_two_layer}. We also consider the setting when $W_p$ is set as $(W\E_{x_1} x_1x_1^\top W^\top)^\alpha$ in Appendix~\ref{sec:diff_corr} and extend the result to deep linear networks in Appendix~\ref{sec:deep_linear}.

Due to the identity initialization, we can ensure that $W$ is always a real symmetric matrix and is simultaneously diagonalizable with $P_B$. We can then analyze the evolution of each eigenvalue in $W$ separately. Under our assumptions, it turns out that all the eigenvalues whose eigenvectors lie in the $B$ subspace share the same value $\lambda_B$, and all the eigenvalues in the $S$ subspace share the value $\lambda_S$ as shown in the following time dynamics: 
\begin{align*}
    &\dot \lambda_B= \lambda_B\br{-(1+\sigma^2)\absr{\lambda_B}^{4\alpha} + \absr{\lambda_B}^{2\alpha} -\eta},\\
    &\dot \lambda_S=  \lambda_S\br{-\absr{\lambda_S}^{4\alpha} + \absr{\lambda_S}^{2\alpha} -\eta}.
\end{align*}
Next, we show $\lambda_B$ converges to zero and $\lambda_S$ converges to a positive number, which immediately implies that $W$ converges to some scaling of $P_S.$

\begin{figure}[t]
\centering
    \includegraphics[width=\linewidth]{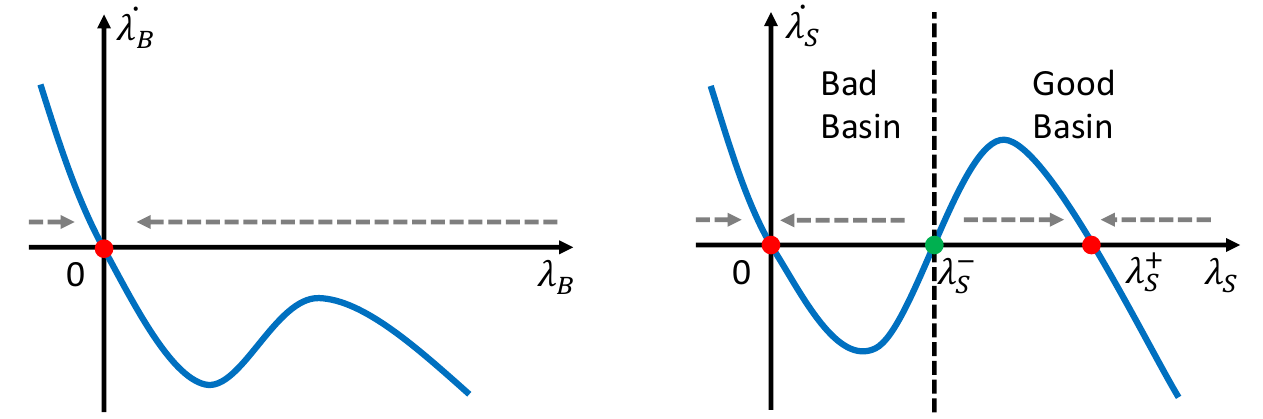}
\caption{\small (\textbf{Left}) with appropriate weight decay, $\lambda_B$ always converges to zero ; (\textbf{Right}) $\lambda_S$ converges to zero when it's initialized in the bad basin and converges to positive $\lambda_S^+$ when it's initialized in the good basin. 
}\label{fig:lambda_dynamics}
\end{figure}

Similar as the analysis in~\citet{tian2021understanding}, when $\eta>\frac{1}{4(1+\sigma^2)},$ we know $\dot \lambda_B <0$ for any $\lambda_B >0$ and $\lambda_B = 0$ is a stable stationary point, as illustrated in Figure~\ref{fig:lambda_dynamics} (\textbf{Left}). Therefore, as long as $\eta>\frac{1}{4(1+\sigma^2)}$, $\lambda_B$ must converge to zero. On the other hand, 
there are three non-negative solutions to $\dot \lambda_S=0$, which are $0, \lambda_S^-=\pr{\frac{1-\sqrt{1-4\eta}}{2}}^{1/(2\alpha)}$ and $\lambda_S^+=\pr{\frac{1+\sqrt{1-4\eta}}{2}}^{1/(2\alpha)}$ when $0<\eta<\frac{1}{4}.$  As illustrated in Figure~\ref{fig:lambda_dynamics} (\textbf{Right}), if initialization $\delta > \lambda_S^-$ (good basin), $\lambda_S$ converges to a positive value $\lambda_S^+;$ if $0<\delta < \lambda_S^-$ (bad basin), $\lambda_S$ converges to zero. 

\paragraph{Thresholding role of weight decay in feature learning: }
While \citet{tian2021understanding} showed why \ncssl does not collapse, one key question is how \ncssl learns useful features and how the method determines which feature is learned. Now it is clear: the weight decay factor $\eta$ makes a call on what features should be learned. As illustrated in Figure~\ref{fig:summary}, \emph{Nuisance features} subject to significant changes under data augmentations have larger variance $(\sigma^2>\frac{1}{4\eta}-1$), the eigenspace corresponding to these features goes to zero; on the other hand, \emph{invariant features} that are robust to data augmentations have much smaller variance $(\sigma^2<\frac{1}{4\eta}-1)$ and these features are kept. In our above analysis, $B$ subspace corresponds to the nuisance features and collapses to zero; $S$ subspace corresponds to the invariant features (whose variance was assumed as zero for simplicity) and is kept after training.

\begin{figure}[t]
\centering
    \includegraphics[width=0.9\linewidth]{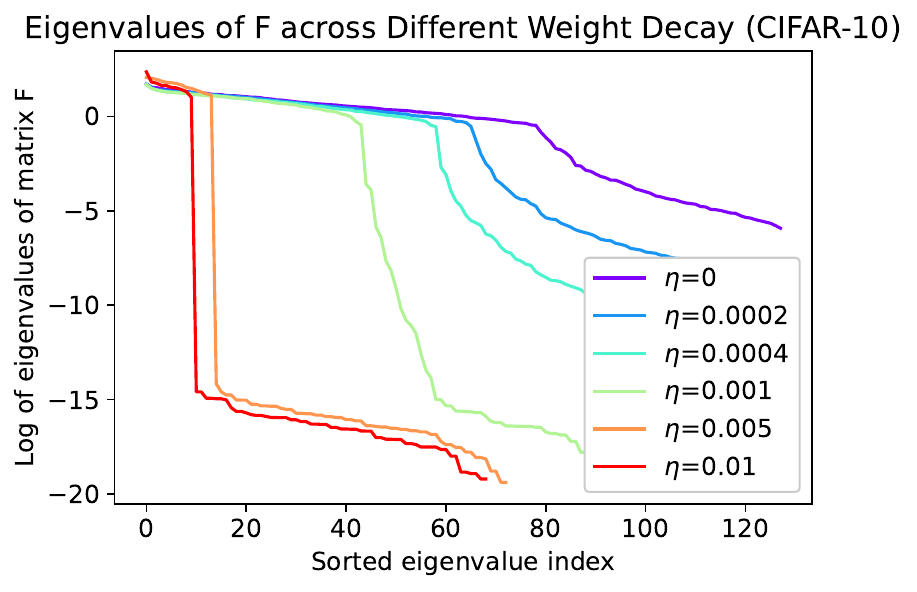}
\caption{\small Eigenvalues of correlation matrix $F$ at $100$-th epoch when it's trained by DirectCopy under different weight decays.
}\label{fig:eigenvalues_weight_decay}
\end{figure}

Figure~\ref{fig:eigenvalues_weight_decay} shows the spectrum of $F$ (which is the correlation matrix of the predictor inputs) when the network is trained by DirectCopy under different weight decay $\eta$ on CIFAR10. The larger the weight decay is, the fewer significant eigenvalues $F$ has~\footnote{Notice that there is a natural drop in the eigenvalues of $F$ even without weight decay $(\eta=0)$ since features along different eigen-directions of $F$ can have very different magnitudes.}. This suggests that the features are better suppressed when larger weight decay is adopted. 

Therefore, it is crucially important to choose weight decay appropriately: a too small $\eta$ may not be sufficient to suppress the nuisance features; a too large $\eta$ can also collapse the invariant features. As shown in Section~\ref{sec:ablation}, both cases lead to worse downstream performance. 

%Interestingly, for empirical experiment, the ``nuisance features'' actually come back after long training, which is beyond what the linear model could predict. We will discuss in Section~\ref{sec:beyond_linear}.  

%We shall see in Section~\ref{sec:ablation} that too small or too large weight decay indeed deteriorate the quality of the learned features and leads to worse performance on downstream tasks. Moreover, 

\subsection{Sample Complexity of \ncssl}
In this subsection, we prove that DirectCopy (one special case of DirectSet($\alpha$) with $\alpha=1$) learns the projection matrix given polynomial number of unlabeled samples.

\begin{restatable}{theorem}{finitesampleGD}\label{thm:finite_sample_GD}
Suppose network architecture and data distribution are as defined in Assumption~\ref{assump:network} and Assumption~\ref{assump:data_distribution}, respectively. Suppose we initialize online network as $\delta I,$ and run DirectCopy on empirical loss (see Eqn.~\ref{eqn:empirical_loss}) with $\stepsize$ step size and $\eta$
weight decay. Assume $\sigma^2 = \Theta(1), \eta \in \pr{ \frac{1+\sigma^2/4}{4(1+\sigma^2)}, \frac{1+3\sigma^2/4}{4(1+\sigma^2)}}, \delta\in (1/2, O(1))$ and $\gamma=\Theta(1).$ 
For any accuracy $\acc>0,$ given $n\geq \poly(d,1/\acc)$ number of samples, with probability at least $0.99$ there exists $t=O(\log(1/\acc))$ such that
$$\n{\twt- \sqrt{\frac{1+\sqrt{1-4\eta}}{2}}P_S}\leq \acc,$$
where $\twt$ is the online network weights at the $t$-th step.
\end{restatable}

The proof proceeds by first proving that gradient descent on the population loss converges in linear rate and then couples the gradient descent dynamics on empirical loss and that on population loss. See the detailed proof in Appendix~\ref{sec:proofs_gd_empirical}.

\subsection{Sample Complexity on Downstream Tasks}
In this subsection, we show that the learned representations can indeed reduce the sample complexity on the downstream tasks. We consider the following data distribution for the down-stream task: 
\begin{assumption}[Downstream data distribution]\label{assump:downstream_data}
Each input $z^{(i)}$ is sampled from $\calN(0, I_d)$ and its label $y^{(i)}=\inner{z^{(i)}}{w^*}+ \xi^{(i)},$ where $w^*$ is the ground truth vector with unit $\ell_2$ norm and $\xi^{(i)}$ is independently sampled from $\calN(0, \beta^2).$ We assume the ground truth $w^*$ lies on an $r$-dimensional subspace $\calS$ and we denote the projection matrix on subspace $\calS$ simply as $P$.
\end{assumption}
In practice, usually the semantically relevant features ($S$ subspace here) are invariant to augmentations and the nuisance features (orthogonal subspace of $S$) have high variance under augmentations. Therefore, by previous analysis, we expect DirectSet($\alpha$) to learn the projection matrix $P.$

Suppose $\{(z^{(i)}, y^{(i)})\}_{i=1}^n$ are $n$ training samples. Each input $z^{(i)}$ is transformed by a matrix $\hP\in\R^{d\times d}$ (for example the learned online network $W$) to get its representation $\hP z^{(i)}.$ The regularized loss is then defined as 
$\hat{L}(w) := \frac{1}{2n}\sum_{i=1}^n\ns{\inner{\hP z^{(i)}}{ w}-y^{(i)}} +\frac{\rho}{2}\ns{w},$ where the regularization coefficient $\rho$ will be chosen carefully to prevent $w$ from overfitting the noise in labels. Note here the regularization has nothing to do with the predictor regularization $\epsilon I$ in DirectSet($\alpha$) algorithm.  

In the below theorem, we show that when $\fn{P-\hP}$ is small, the above ridge regression can recover the ground truth $w^*$ given only $O(r)$ number of samples, where $r$ is the dimension of the subspace on which $w^*$ lies. 

\begin{restatable}{theorem}{downstream}\label{thm:downstream}
Suppose the downstream data distribution is as defined in Assumption~\ref{assump:downstream_data}.
Suppose $\fn{\hP-P}\leq \acc$ with $\acc<1.$ Choose the regularizer coefficient $\rho = \acc^{1/3}$. For any $\zeta<1/2,$ given $n\geq O(r+\log(1/\zeta))$ number of samples, with probability at least $1-\zeta,$ the training loss minimizer $\hw$ satisfies 
$$\n{\hP \hw - w^* }\leq O\pr{\acc^{1/3} + \beta\frac{\sqrt{r} + \sqrt{\log(1/\zeta)} }{\sqrt{n}}}.$$
\end{restatable}

In the above theorem, when $n$ is at least $O\pr{\frac{\beta^2\pr{r+\log(1/\zeta)}}{\acc^{2/3}}},$ we have $\n{\hP \hw - w^* }\leq O(\acc^{1/3}).$ Note that if we directly estimate $\hw$ without transforming the inputs by $\hP$, we need $\Omega(d)$ number of samples to ensure that $\n{\hw - w^*}\leq o(1)$~\citep{wainwright2019high}.
The proof of Theorem~\ref{thm:downstream}
follows from bounding the difference between $\hP \hw$ and $w^*$ by matrix concentration inequalities and matrix perturbation bounds. The full proof is in Appendix~\ref{sec:proofs_down_stream}.
%\vspace{-0.1in}
\section{Empirical Performance of DirectCopy}
%\vspace{-0.1in}
\label{sec:empirical_perf}
In the previous analysis, we show DirectSet($\alpha$), and in particular DirectCopy (DirectSet($\alpha$) with $\alpha=1$), could recover the input feature structure with polynomial samples and make the downstream task more sample efficient in a simple linear setting. Compared with the original DirectPred (DirectSet($\alpha$) with $\alpha=1/2$), DirectCopy is a simpler and computationally more efficient algorithm since it directly set the predictor as the correlation matrix $F$, without the eigen-decomposition step. By our analysis in Theorem~\ref{thm:learn_subspace}, DirectCopy also learns the projection matrix $P_S$ with larger scale~\footnote{Recall that in Theorem~\ref{thm:learn_subspace} under DirectSet($\alpha$), online matrix $W$ converges to $\pr{\frac{1+\sqrt{1-4\eta}}{2}}^{1/(2\alpha)}P_S$. So with a larger $\alpha$, the scalar in front of $P_S$ becomes larger.} compared with DirectPred, which suggests that the invariant features learned by DirectCopy are stronger and more distinguishable. Next, we show that DirectCopy is on par with (or even outperforms) the original DirectPred in various datasets, when coupling with deep nonlinear models on real datasets.  

%theoretical properties. 
%The analysis in the last section showed that in a simple linear setting, DirectSet($\alpha$) with any positive $\alpha$ can learn a useful projection matrix. 

%Motivated by this analysis, we propose DirectCopy algorithm (or DirectSet(1)) that avoids computing the eigen-decomposition of matrix $F$.  
%\vspace{-0.1cm}
\subsection{Results on STL-10, CIFAR-10 and CIFAR-100}
%\vspace{-0.1cm}
We use ResNet-18~\citep{resnet} as the backbone network, a two-layer nonlinear MLP as the projector, and a linear predictor. Unless specified otherwise, SGD is used as the optimizer with weight decay $\eta = 0.0004$. To evaluate the quality of the pre-trained representations, we follow the linear evaluation protocol. Each setting is repeated 5 times to compute the mean and standard deviation. The accuracy is reported as ``mean$\pm$std''. Unless explicitly specified, we use learning rate $\stepsize=0.01,$ regularization $\epsilon=0.2$ on STL-10; $\stepsize=0.02, \epsilon=0.3$ on CIFAR-10 and $\stepsize=0.03, \epsilon=0.3$ on CIFAR-100. See more detailed experiment settings in Appendix~\ref{sec:experiment_settings}.

\textbf{STL-10: } We evaluate the quality of the learned representation after each epoch, and report the best accuracy in the first 100/300/500 epochs in Table~\ref{tbl:stl10_cifar10}. DirectCopy achieves substantially better performance than DirectPred and SGD baseline, especially when trained with longer epochs. 
DirectPred (freq=5) means the predictor is set by DirectPred every 5 batchs, and is trained with gradient updates in other batchs, which outperforms DirectPred in later epochs, but is still much worse than DirectCopy. The SGD baseline is obtained by training the linear predictor using SGD.
%On STL-10, DirectCopy is run with learning rate 0.01 and regularization $\epsilon=0.2.$ 

\textbf{CIFAR-10/100:} For CIFAR-10, DirectCopy is slighly worse than DirectPred at epoch 100, but catches up and gets even better performance in epoch 300 and 500 (Table~\ref{tbl:stl10_cifar10}). %On CIFAR-10, DirectCopy is trained with learning rate 0.02 and regularization parameter $\epsilon=0.3.$
For CIFAR-100, at earlier epochs, the performance of DirectCopy is not as good as DirectPred, but the gap gradually diminishes in later epochs. Both DirectCopy and DirectPred outperfom the SGD baseline. DirectPred (freq=5) achieves even better performance, but at the cost of a more complicated algorithm.

%On CIFAR-100, both DirectCopy and DirectPred are trained with step size 0.03 and regularization $\epsilon=0.3.$

\begin{table}[t]
\caption{\small STL-10/CIFAR-10/CIFAR-100 Top-1 accuracy of DirectCopy and other algorithms. The numbers for DirectPred, DirectPred (freq=5) and SGD baseline on STL-10/CIFAR-10 are obtained from~\citet{tian2021understanding}.}
\vskip 0.15in
\centering
\small
\setlength{\tabcolsep}{1pt}
    \begin{tabular}{c||c|c|c}
    & \multicolumn{3}{c}{Num of epochs}\\
                        &   100 & 300 & 500 \\
     \hline 
     \multicolumn{4}{c}{\emph{STL-10}} \\
     \hline
     \small DirectCopy   & $77.83{\pm}0.56$  & $\mathbf{82.01{\pm}0.28}$  & $\mathbf{82.95 {\pm}0.29}$ \\
     DirectPred   & $\mathbf{77.86{\pm}0.16}$  & $78.77{\pm}0.97$  & $78.86 {\pm}1.15$ \\ 
     DirectPred (freq=5) & $77.54 {\pm}0.11$  & $79.90{\pm}0.66$  & $80.28{\pm}0.62$ \\
     SGD baseline  &    $75.06{\pm}0.52$  & $75.25 {\pm}0.74$  &  $75.25 {\pm}0.74$ \\
     \hline 
     \multicolumn{4}{c}{\emph{CIFAR-10}} \\
     \hline
     DirectCopy          & $84.02{\pm}0.37$  & $\mathbf{89.17{\pm}0.12}$ & $\mathbf{89.62{\pm}0.10}$ \\
    DirectPred        & $\mathbf{85.21{\pm}0.23}$  & $88.88{\pm}0.15$ & $89.52{\pm}0.04$ \\
    DirectPred (freq=5) & $84.93{\pm}0.29$ & $88.83{\pm}0.10$ & $89.56{\pm}0.13$ \\
    SGD baseline          & $84.49{\pm}0.20$ & $88.57{\pm}0.15$ & $89.33{\pm}0.27$\\
    \hline 
     \multicolumn{4}{c}{\emph{CIFAR-100}} \\
     \hline
     DirectCopy          & $55.40{\pm}0.19$  & $61.06{\pm}0.14$ & $62.23{\pm}0.06$ \\
    DirectPred        & $\mathbf{56.60{\pm}0.27}$  & $61.65{\pm}0.18$ & $62.68{\pm}0.35$\\ 
    DirectPred (freq=5) & $56.43{\pm}0.21$ & $\mathbf{62.01{\pm}0.22}$ & $\mathbf{63.15{\pm}0.27}$ \\
    SGD baseline          & $54.94{\pm}0.50$ & $60.88{\pm}0.59$ & $61.42{\pm}0.89$\\
     \end{tabular}
    \vskip -0.1in
    
    \label{tbl:stl10_cifar10}
    
\end{table}

\subsection{Results on ImageNet}
\vskip -0.1in
\begin{table}[h]
\caption{\small ImageNet Top-1 accuracy of DirectCopy, DirectPred and BYOL baseline with one/two-layer predictor after 100 epochs.}
\vskip 0.15in
\centering
\small
\setlength{\tabcolsep}{1pt}
\begin{tabular}{c||c|c|c|c}
     
     &   DirectCopy & DirectPred & 1-layer BYOL & 2-layer BYOL\\
     \hline
     ImageNet & $\mathbf{68.8}$ & $68.5$ & $68.6$ & $66.5$
     \end{tabular}

    \label{tbl:imagenet}
    \vskip -0.1in

\end{table}
%Following BYOL~\citep{byol}, we use ResNet-50 as the backbone and a two-layer MLP as the projector. We use LARS~\citep{You2017} optimizer and trains the model for 100 epochs. See the detailed experiment setting in Appendix~\ref{sec:experiment_settings}.

Following BYOL~\citep{byol}, we use ResNet-50 as the backbone and a two-layer MLP as the projector. We use LARS~\citep{You2017} optimizer and train the model for 100 epochs. See more detailed experiment settings in Appendix~\ref{sec:experiment_settings}.

For fairness, we compare \ours to the gradient-based baseline which uses the same-sized linear predictor as ours. As shown in Table~\ref{tbl:imagenet}, at 100-epoch, this baseline achieves 68.6 top-1 accuracy, which is already significantly higher than BYOL with two-layer predictor reported in the literature (e.g., \citet{chen2020exploring} reported 66.5 top-1 under 100-epoch training). \ours using normalized $F$ with regularization parameter $\epsilon = 0.01$ achieves 68.8 under the same setting, better than this strong baseline. In contrast, DirectPred achieves 68.5, slightly lower than the BYOL baseline with linear predictor.

\section{Ablation Study}
\label{sec:ablation}
In this section, we study the influence of predictor regularization $\epsilon$, normalization method, weight decay and degree $\alpha$ on the performance of DirectCopy.

\begin{table}[h]
\caption{\small STL-10/CIFAR-10 Top-1 accuracy of DirectCopy with varying regularization $\epsilon$.}
\vskip 0.15in
\centering
\small
\begin{tabular}{c||c|c}
    & \multicolumn{2}{c}{Number of epochs}\\
                        &   100 & 300 \\
     \hline 
     \multicolumn{3}{c}{\emph{STL-10}} \\
     \hline
     $\epsilon=0$   & $76.57{\pm}0.66$  & $81.19{\pm}0.39$\\
     $\epsilon=0.1$   & $\mathbf{78.05{\pm}0.14}$  & $81.60{\pm}0.15$\\
     $\epsilon=0.2$   & $77.83{\pm}0.56$  & $\mathbf{82.01{\pm}0.28}$\\
     $\epsilon=1$   & $31.10{\pm}0.80$  & $31.10{\pm}0.80$\\
     \hline 
     \multicolumn{3}{c}{\emph{CIFAR-10}} \\
     \hline
     $\epsilon=0$   & $80.53{\pm}1.14$  & $86.07{\pm}0.71$\\
     $\epsilon=0.1$   & $83.97{\pm}0.25$  & $88.58{\pm}0.11$\\
     $\epsilon=0.3$   & $\mathbf{84.02{\pm}0.37}$  & $\mathbf{89.17{\pm}0.12}$\\
     $\epsilon=1$   & $57.38{\pm}11.62$  & $83.15{\pm}4.24$\\
     \end{tabular}
    
    \label{tbl:epsilon}
\vskip -0.1in
\end{table}

\paragraph{Predictor regularization: } Table~\ref{tbl:epsilon} shows that when the predictor regularization $\epsilon$ increases, the performance of DirectCopy on STL-10 and CIFAR-10 improves at first and then deteriorates. On STL-10, DirectCopy with $\epsilon=1$ completely fails. On CIFAR-10, although DirectCopy with $\epsilon=1$ achieved reasonable performance at epoch 300, it's still much worse than $\epsilon=0.3.$

\begin{figure}[h]
\centering
    \includegraphics[width=\linewidth]{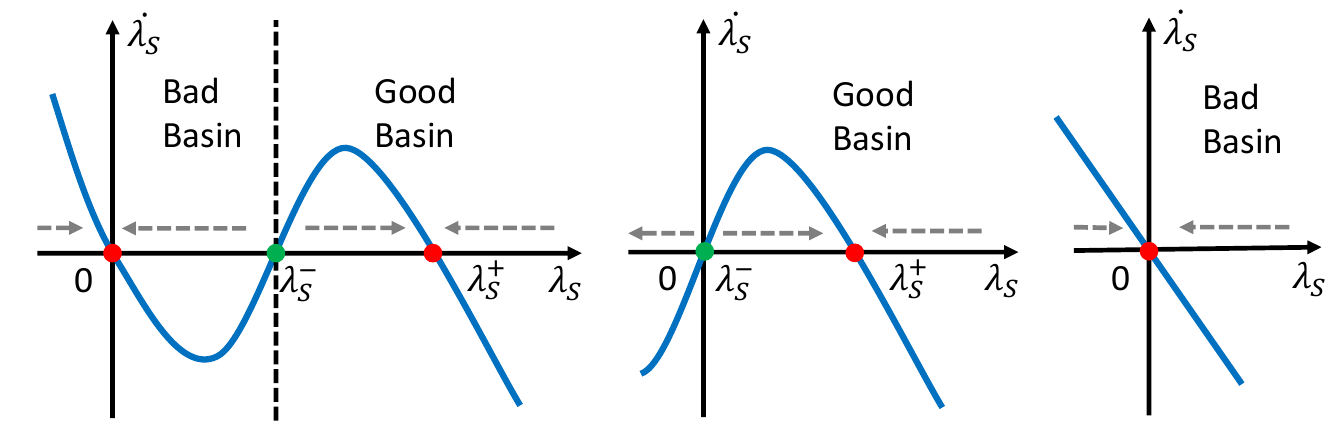}

\caption{\small Increasing $\epsilon$ shifts the two positive stationary points $\lambda_S^-$ and $\lambda_S^+$ towards zero. (\textbf{Left}) when $0\leq \epsilon<\frac{1-\sqrt{1-4\eta}}{2},$ increasing $\epsilon$ expands the good basin ($\lambda_S>\lambda_S^-$) by reducing $\lambda_S^-$. (\textbf{Middle}) when $\frac{1-\sqrt{1-4\eta}}{2}\leq \epsilon<\frac{1+\sqrt{1-4\eta}}{2},$ $\lambda_S^-$ becomes zero and $\lambda_S$ converges to positive $\lambda_S^+$ from any positive value; further increasing $\epsilon$ decreases $\lambda_S^+$. (\textbf{Right}) when $\frac{1+\sqrt{1-4\eta}}{2}\leq \epsilon,$  $\lambda_S^+$ becomes zero and $\lambda_S$ always converges to zero.}
\label{fig:lambda_dynamics_eps}

\end{figure}

To better understand the role of $\epsilon,$ we analyze the simple linear setting as in Section~\ref{sec:two_layer_setup} while setting $W_p = WW^\top +\epsilon I.$ Recall that $\lambda_B$ is the eigenvalue of $W$ in $B$ subspace and $\lambda_S$ is that in $S$ subspace. When the weight decay is appropriate, $\lambda_B$ still converges to zero. On the other hand, the dynamics for $\lambda_S$ is as follows:
$
    \dot \lambda_S 
    = -\lambda_S\pr{\lambda_S^2+\epsilon-\frac{1-\sqrt{1-4\eta}}{2}}\pr{\lambda_S^2+\epsilon-\frac{1+\sqrt{1-4\eta}}{2}}.
$
Increasing $\epsilon$ shifts the two positive stationary points $\lambda_S^-,\lambda_S^+$ towards zero. As illustrated in Figure~\ref{fig:lambda_dynamics_eps}, as $\epsilon$ increases, when $\lambda_S^+$ is still positive, the good attraction basin expands, which means $\lambda_S$ can converge to a positive value from a smaller initialization; when $\lambda_S^+$ shifts to zero, $\lambda_S$ converges to zero regardless the initialization size. See the full analysis in Appendix~\ref{sec:proofs_eps_identity}.

\begin{figure}[h]
\vskip -0.15in
\centering
    \includegraphics[width=0.9\linewidth]{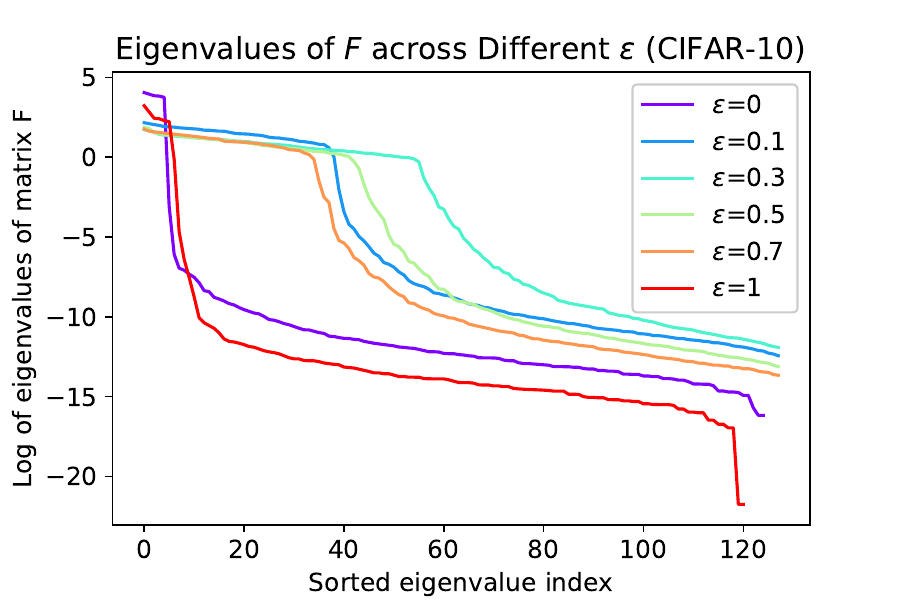}

\caption{\small Eigenvalues of $F$ when trained by DirectCopy under different predictor regularization $\epsilon$ on CIFAR-10 for 100 epochs. }\label{fig:change_of_eps}
\vskip -0.1in
\end{figure}

Intuitively, a reasonable $\epsilon$ can alleviate representation collapse, but a too large $\epsilon$ also encourages representation collapse. As shown in Figure~\ref{fig:change_of_eps}, when $\epsilon$ increases from zero, more eigenvalues of $F$ becomes large; but when $\epsilon$ exceeds $0.3$, eigenvalues of $F$ begin to collapse. 

\begin{table}[h]
\caption{\small STL-10/CIFAR-10 Top-1 accuracy of DirectCopy with $F$ matrix normalized by spectral norm/Frobenius norm or no normalization.}
\vskip 0.15in
\centering
\small
\begin{tabular}{c||c|c}
    & \multicolumn{2}{c}{Number of epochs}\\
                        &   100 & 300 \\
     \hline 
     \multicolumn{3}{c}{\emph{STL-10}} \\
     \hline
     Spectral   & $\mathbf{77.83{\pm}0.56}$  & $82.01{\pm}0.28$\\
     Frobenius   & $77.71{\pm}0.18$  & $\mathbf{82.06{\pm}0.28}$\\
     None   & $77.81{\pm}0.20$  & $82.00{\pm}1.24$\\
     \hline 
     \multicolumn{3}{c}{\emph{CIFAR-10}} \\
     \hline
     Spectral   & $84.02{\pm}0.37$  & $89.17{\pm}0.12$\\
     Frobenius   & $\mathbf{84.33{\pm}0.25}$  & $\mathbf{89.62{\pm}0.14}$\\
     None   & $81.76{\pm}0.34$  & $89.21{\pm}0.17$\\
     \end{tabular}
    \label{tbl:normalization}
\vskip -0.1in
\end{table}

\paragraph{Normalization on $F$: } In our experiments, we have been normalizing $F$ by its spectral norm before adding the regularization: $W_p = F/\n{F} +\epsilon I.$ It turns out that we can also normalize $F$ by its Frobenius norm or simply skip the normalization step. In Table~\ref{tbl:normalization}, we see comparable performance from DirectCopy with Frobenius normalization or no normalization, especially when trained longer.

\begin{table}[h]
\caption{\small STL-10/CIFAR-10 Top-1 accuracy of DirectCopy with varying weight decay.}
\vskip 0.15in
\centering
\small
\begin{tabular}{c||c|c}
    & \multicolumn{2}{c}{Number of epochs}\\
                        &   100 & 300 \\
     \hline 
     \multicolumn{3}{c}{\emph{STL-10}} \\
     \hline
     $\eta=0$   & $71.94{\pm}0.93$  & $78.53{\pm}0.40$\\
     $\eta=0.0004$   & $\mathbf{77.83{\pm}0.56}$  & $\mathbf{82.01{\pm}0.28}$\\
     $\eta=0.001$   & $77.65{\pm}0.16$  & $80.28{\pm}0.16$\\
     $\eta=0.01$   & $58.12{\pm}0.94$  & $58.53{\pm}0.76$\\
     \hline 
     \multicolumn{3}{c}{\emph{CIFAR-10}} \\
     \hline
     $\eta=0$   & $79.15{\pm}0.08$  & $85.35{\pm}0.31$\\
     $\eta=0.0004$   & $\mathbf{84.02{\pm}0.37}$  & $\mathbf{89.17{\pm}0.12}$\\
     $\eta=0.001$   & $83.91{\pm}0.33$  & $87.75{\pm}0.16$\\
     $\eta=0.01$   & $65.31{\pm}1.19$  & $65.63{\pm}1.30$\\
     \end{tabular}

    \label{tbl:weight_decay}
\vskip -0.1in
\end{table}

\paragraph{Weight decay: } Table~\ref{tbl:weight_decay} shows that when weight decay $\eta$ increases, the performance of DirectCopy improves at first and then deteriorates. This fits our analysis on simple linear networks. Basically, when the weight decay $\eta$ increases, it can suppress the nuisance features more effectively, but a too large weight decay also collapses the useful features.

\paragraph{Predictor degree:} We compare DirectCopy against DirectSet($\alpha$) with $\alpha=2, 1/2,1/4.$ Table~\ref{tbl:alpha} shows that DirectCopy outperforms other algorithms on STL-10. On CIFAR-10, DirectCopy is slightly worse at epoch 100, but catches up in later epochs. According to our analysis, $\alpha=2$ is supposed to learn stronger invariant features than $\alpha=1$, but it does not lead to better performance in experiments. This suggests that the benefits from more distinguishable features diminish beyond $\alpha=1$. 

\begin{table}[h]
\caption{\small STL-10/CIFAR-10 Top-1 accuracy of DirectSet($\alpha$) with varying degree $\alpha$.}
\vskip 0.15in
\centering
\small
\begin{tabular}{c||c|c}
    & \multicolumn{2}{c}{Number of epochs}\\
                        &   100 & 300 \\
     \hline 
     \multicolumn{3}{c}{\emph{STL-10}} \\
     \hline
     $\alpha=2$   & $76.80{\pm}0.22$  & $80.90{\pm}0.18$\\
     $\alpha=1$   & $\mathbf{77.83{\pm}0.56}$  & $\mathbf{82.01{\pm}0.28}$\\
     $\alpha=1/2$   & $77.82{\pm}0.37$  & $77.83{\pm}0.37$\\
     $\alpha=1/4$   & $76.82{\pm}0.36$  & $76.82{\pm}0.36$\\
     \hline 
     \multicolumn{3}{c}{\emph{CIFAR-10}} \\
     \hline
     $\alpha=2$   & $82.96{\pm}0.56$  & $88.60{\pm}0.11$\\
     $\alpha=1$   & $84.02{\pm}0.37$  & $\mathbf{89.17{\pm}0.12}$\\
     $\alpha=1/2$   & $\mathbf{84.88{\pm}0.21}$  & $88.32{\pm}0.57$\\
     $\alpha=1/4$   & $84.78{\pm}0.21$  & $87.82{\pm}0.32$\\
     \end{tabular}
    \label{tbl:alpha}
    \vskip -0.1in
\end{table}

\section{Conclusion}
\label{sec:conclusion}

In this paper, we have proved DirectSet($\alpha$) can learn the desirable projection matrix in a linear network setting and can reduce the sample complexity on down-stream tasks. Our analysis sheds light on the crucial role of weight decay in \ncssl, which discards the features that have high variance under augmentations and keeps the invariant features. Inspired by the analysis, we designed a simpler and more efficient algorithm DirectCopy, which achieved comparable or even better performance than the original DirectPred~\citep{tian2021understanding} on various datasets.

We view our paper as an initial step towards demystifying the representation learning in \ncssl. Many mysteries still lie beyond the explanation of the current theory and we leave them for future work.  

%\section*{Reproducibility Statement}
%For our theory, the assumptions and theorems have been clearly stated in the main paper, and the complete proof of all claims can be found in the Appendix. For the experiments, we described the detailed settings in Appendix~\ref{sec:experiment_settings}.

\bibliography{ref}
\bibliographystyle{icml2022}

\newpage
\appendix
\onecolumn
\section{Detailed Experiment Setting}\label{sec:experiment_settings}
\paragraph{STL-10, CIFAR-10, CIFAR-100}: 
We use ResNet-18~\citep{resnet} as the backbone network, a two-layer nonlinear MLP (with batch normalization, ReLU activation, hidden layer width 512, output width 128) as the projector, and a linear predictor. Unless specified otherwise, SGD is used as the optimizer with momentum $0.9$, weight decay $\eta = 0.0004$ and batch size 128. The EMA parameter for the target network is set as 0.996 and the EMA parameter $\mu$ of the correlation matrix $\hat{F}$ is set as 0.5. Our code is adapted from~\cite{tian2021understanding}~\footnote{Their open source code is at \href{ https://github.com/facebookresearch/luckmatters/tree/main/ssl}{https://github.com/facebookresearch/luckmatters/tree/main/ssl}}, and we follow the same data augmentation process. 

To evaluate the quality of the pre-trained representations, we follow the linear evaluation protocol. Each setting is repeated 5 times to compute the mean and standard deviation. The accuracy is reported as ``mean$\pm$std''. Unless explicitly specified, we use learning rate $\stepsize=0.01,$ regularization $\epsilon=0.2$ on STL-10; $\stepsize=0.02, \epsilon=0.3$ on CIFAR-10 and $\stepsize=0.03, \epsilon=0.3$ on CIFAR-100.

\paragraph{ImageNet}:
Following BYOL~\citep{byol}, we use ResNet-50 as the backbone and a two-layer MLP (with batch normalization, ReLU, hidden layer width 4096, output width 256) as the projector. We use LARS~\citep{You2017} optimizer and trains the model for 100 epochs, with a batch size 4096. The learning rate is 7.2, which is linearly scaled from the base learning rate 0.45 at batch size 256. Other setups such as weight decay ($\eta = 1e^{-6}$), target EMA (scheduled from 0.99 to 1), augmentation recipe (color jitters, blur, etc.), and linear evaluation protocol are the same as BYOL.

\iffalse
\subsection{Phenomena Beyond Linear Models}\label{sec:beyond_linear_experiments}
\begin{figure}[h]
\centering
    \includegraphics[width=\linewidth]{figures/features_cdf.pdf}
\caption{\small Comparison of features learned by DirectCopy on CIFAR-10 under different weight decay. The left two figures are for features after the projector and the right two are for features before the projector. Each feature vector is a 4096-dimensional vector that consists of the values of one output neuron of the trained online network (before or after projector) across 4096 training samples. We normalize each feature vector to mean zero and unit $\ell_2$ norm. The correlation between two feature vector $f_1,f_2$ is defined as $\absr{\inner{f_1}{f_2}}.$ The correlation between a feature vector and a model is defined as the best correlation between this feature and all the features in that model. In each figure, CDF$[a,b](x)$ denotes the fraction of features in the model with $\eta=a$ that has correlation at most $x$ with the other model with $\eta=b.$
}\label{fig:features_cdf}
\end{figure}
\fi
\section{Proofs of Single-layer Linear Networks}

\subsection{Gradient Flow on Population Loss}\label{sec:proofs_two_layer}
In this section, we give the proof of Theorem~\ref{thm:learn_subspace}, which shows that DirectSet($\alpha$) running on the population loss with infinitesimal learning rate and $\eta$ weight decay can learn the projection matrix onto subspace $S$.

\learnsubspace*

As we already mentioned in the main text, Theorem~\ref{thm:learn_subspace} is proved by analyzing each eigenvalue of $W$ separately. We show that the eigenvalues in the $B$ subspace converge to zero, and the eigenvalues in the $S$ subspace converge to the same positive number, which immediately implies that $W$ converges to a scaling of the projection matrix $P_S.$

\begin{proofof}{Theorem~\ref{thm:learn_subspace}}
We can compute the gradient in terms of $W$ as follows,
\begin{align*}
\nabla L(W) 
=& \E_{x_1, x_2} W_p^\top\pr{W_p W x_1 - W_a x_2 } x_1^\top\\
=& W_p^\top\pr{W_p W \E_{x_1} x_1 x_1^\top  - W_a \E_{x_1, x_2} x_2 x_1^\top }.
\end{align*}
Note that the two augmented views $x_1, x_2$ are sampled by first sampling input $x$ from $\N(0,I_d)$, and then independently sampling $x_1, x_2$ from $\N(x, \sigma^2P_B).$ Therefore, we know $\E_{x_1} x_1 x_1^\top = I+\sigma^2 P_B$ and $\E_{x_1, x_2} x_2 x_1^\top=I.$ Recall that we run gradient flow on $W$ with weight decay $\eta,$ so the dynamics on $W$ is as follows:
\begin{align*}
    \dot W = & W_p^\top (-W_p W(I+\sigma^2 P_B) + W_a) - \eta W,
\end{align*}
where the first term comes from the gradient and the second term is due to weight decay. 

Since $W$ is initialized as $\delta I,$ and $W_a = W, W_p=(WW^\top)^\alpha,$ so we know initially $W,W_p,W_a, I$ and $P_B$ are all simultaneously diagonalizable, which then implies $\dot{W}$ is simultaneously diagonalizable with $W$. This argument can continue to show that at any time point, $W,W_p,W_a, I$ and $P_B$ are all simultaneously diagonalizable. Since $W$ is always a real symmetric matrix, we have $W_p =(WW^\top)^\alpha =\absr{W}^{2\alpha}.$ The dynamics on $W$ can then be written as 
\begin{align*}
    \dot W = & \absr{W}^{2\alpha} (-\absr{W}^{2\alpha} W(I+\sigma^2 P_B) + W) - \eta W\\
    =& W\pr{-(I+\sigma^2 P_B) \absr{W}^{4\alpha} + \absr{W}^{2\alpha} -\eta}.
\end{align*}

Let the eigenvalue decomposition of $W$ be $\sum_{i=1}^d \lambda_i u_i u_i^\top,$ with span$(\{u_{d-r+1}, \cdots, u_d\})$ equals to subspace $B$. We can separately analyze the dynamics of each $\lambda_i.$ Furthermore, we know $\lambda_1, \cdots, \lambda_r$ have the same value $\lambda_S$ and $\lambda_{d-r+1},\cdots, \lambda_d$ have the same value $\lambda_B.$
Next, we separately show that $\lambda_B$ converge to zero and $\lambda_S$ converges to a positive value.

\paragraph{Dynamics for $\lambda_B$:} We can write down the dynamics for $\lambda_B$ as follows:
$$
\dot \lambda_B = \lambda_B\br{-(1+\sigma^2)\absr{\lambda_B}^{4\alpha} + \absr{\lambda_B}^{2\alpha} -\eta}
$$
Similar as the analysis in~\cite{tian2021understanding}, when $\eta>\frac{1}{4(1+\sigma^2)},$ we know $\dot \lambda_B <0$ for any $\lambda_B >0$ and $\lambda_B = 0$ is a critical point. This means, as long as $\eta>\frac{1}{4(1+\sigma^2)}$, $\lambda_B$ must converge to zero.

\paragraph{Dynamics for $\lambda_S$:} We can write down the dynamics for $\lambda_S$ as follows:
$$
\dot \lambda_S = \lambda_S\br{-\absr{\lambda_S}^{4\alpha} + \absr{\lambda_S}^{2\alpha} -\eta}.
$$
When $0<\eta<\frac{1}{4},$ we know $\dot \lambda_S>0$ for $\lambda_S^{2\alpha}\in \pr{\frac{1-\sqrt{1-4\eta}}{2}, \frac{1+\sqrt{1-4\eta}}{2}}$ and $\dot \lambda_S<0$ for $\lambda_S^{2\alpha}\in \pr{ \frac{1+\sqrt{1-4\eta}}{2}, \infty}.$ Furthermore, we know $\dot \lambda_S=0$ when $\lambda_S^{2\alpha} = \frac{1+\sqrt{1-4\eta}}{2}.$ Therefore, as long as $0<\eta<\frac{1}{4}$ and initialization $\delta^{2\alpha} > \frac{1-\sqrt{1-4\eta}}{2}$, we know $\lambda_S^{2\alpha}$ converges to $\frac{1+\sqrt{1-4\eta}}{2}.$

Overall, we know when $\frac{1}{4(1+\sigma^2)}<\eta<\frac{1}{4}$ and $\delta>\pr{\frac{1-\sqrt{1-4\eta}}{2}}^{1/(2\alpha)},$ we have $\lambda_B$ converge to zero and $\lambda_S$ converge to $\pr{\frac{1+\sqrt{1-4\eta}}{2}}^{1/(2\alpha)}.$ That is, matrix $W$ converges to $\pr{\frac{1+\sqrt{1-4\eta}}{2}}^{1/(2\alpha)}P_S.$
\end{proofof}

\subsection{Gradient Descent on Empirical Loss}\label{sec:proofs_gd_empirical}
In this section, we prove that DirectCopy successfully learns the projection matrix given polynomial number of samples.

\finitesampleGD*

When running gradient descent on the empirical loss, the eigenspace of $\twt$ can shift and become no longer simultaneously diagonalizable with $P_B$. So we cannot independently analyze each eigenvalue of $\twt$ as before, which brings significant challenge into the analysis. Instead of directly analyzing the dynamics of $\twt,$ we first show that the gradient descent iterates $W_t$ on the population loss converges to $P_S$ in linear rate, and then show that $\twt$ stays close to $\wt$ within certain iterations.

\begin{lemma}\label{lem:population_GD}
In the setting of Theorem~\ref{thm:finite_sample_GD}, let ${W_t}$ be the gradient descent iterations on the population loss $L$. Given any accuracy $\acc>0,$ for any $t\geq C\log(1/\acc),$ we have 
$$\n{W_t- \sqrt{\frac{1+\sqrt{1-4\eta}}{2}}P_S}\leq \acc,$$
where $C$ is a positive constant. 
\end{lemma}

The proof of Lemma~\ref{lem:population_GD} is similar as the gradient flow analysis in Section~\ref{sec:GF_population}. 
Next, we show that the gradient descent trajectory on the empirical loss stays close to the gradient descent trajectory on the population loss within $O(\log(1/\acc))$ iterations.

\begin{lemma}\label{lem:coupling}
In the setting of Theorem~\ref{thm:finite_sample_GD}, let ${W_t}$ be the gradient descent iterations on the population loss and let ${\twt}$ be the gradient descent iterations on the empirical loss. For any accuracy $\acc>0,$ given $n\geq \poly(d,1/\acc)$ number of samples, with probability at least $0.99$, for any $t\leq C\log(1/\acc),$ we have
$$\n{\twt- \wt}\leq \acc,$$
where the constant $C$ comes from Lemma~\ref{lem:population_GD}.
\end{lemma}
Then the proof of Theorem~\ref{thm:finite_sample_GD} directly follows from Lemma~\ref{lem:population_GD} and Lemma~\ref{lem:coupling}.

\begin{proofof}{Theorem~\ref{thm:finite_sample_GD}}
According to Lemma~\ref{lem:population_GD}, we know given any accuracy $\acc',$ for $t=C\log(1/\acc),$ we have 
$$\n{W_t- \sqrt{\frac{1+\sqrt{1-4\eta}}{2}}P_S}\leq \acc',$$
where $C$ is a positive constant. 

According to Lemma~\ref{lem:coupling}, we know given $n\geq \poly(d,1/\acc')$ number of samples, with probability at least $0.99$, 
$$\n{\twt- \wt}\leq \acc'.$$

Therefore, we have 
$$\n{\twt- \sqrt{\frac{1+\sqrt{1-4\eta}}{2}}P_S}\leq \n{\wt- \sqrt{\frac{1+\sqrt{1-4\eta}}{2}}P_S}+\n{\twt- \wt}\leq 2\acc'.$$
Replacing $\acc'$ by $\acc/2$ finishes the proof.
\end{proofof}

In section~\ref{sec:lemma_proof}, we give the proof of Lemma~\ref{lem:population_GD} and Lemma~\ref{lem:coupling}. Proofs of some technical lemmas are left in Appendix~\ref{sec:technical}.

\subsubsection{Proofs for Lemma~\ref{lem:population_GD} and Lemma~\ref{lem:coupling}}\label{sec:lemma_proof}

\begin{proofof}{Lemma~\ref{lem:population_GD}}
Similar as in Theorem~\ref{thm:learn_subspace}, we can show that at any step $t$, $\wt$ is simultaneously diagonalizable with $\wat, \wpt, I$ and $P_B.$ The update on $\wt$ is as follows,
\begin{align*}
    W_{t+1} =  \wt + \gamma\wt\pr{-(I+\sigma^2 P_B) \wt^4 + \wt^2 -\eta}.
\end{align*}

Let the eigenvalue decomposition of $\wt$ be $\sum_{i=1}^d \lambda_{i,t} u_i u_i^\top,$ with span$(\{u_{d-r+1}, \cdots, u_d\})$ equals to subspace $B$. We can separately analyze the dynamics of each $\lambda_{i,t}.$ Furthermore, we know $\lambda_{1,t}, \cdots, \lambda_{r,t}$ have the same value $\lambda_{S,t}$ and $\lambda_{d-r+1,t},\cdots, \lambda_{d,t}$ have the same value $\lambda_{B,t}.$
Next, we separately show that $\lambda_{B,t}$ converge to zero and $\lambda_{S,t}$ converges to a positive value in linear rate. 

\paragraph{Dynamics of $\lambda_{B,t}$:} We show that 
$$0\leq \lambda_{B,t}\leq (1-\stepsize C_1)^t \delta$$
for any step size $\stepsize\leq C_2,$ where $C_1,C_2$ are two positive constants. 

According to the gradient update, we have 
$$\lambda_{B,t+1} = \lambda_{B,t}+\stepsize \lambda_{B,t}\br{-(1+\sigma^2)\lambda_{B,t}^4 + \lambda_{B,t}^2 -\eta}.$$
We only need to prove that for any $\lambda_{B,t}\in [0,\delta],$ we have 
$$-(1+\sigma^2)\lambda_{B,t}^4 + \lambda_{B,t}^2 -\eta = -\Theta(1).$$
This is true since $\eta \in \pr{ \frac{1+\sigma^2/4}{4(1+\sigma^2)}, \frac{1+3\sigma^2/4}{4(1+\sigma^2)}}$ and $\sigma^2,\delta$ are two positive constants.

\paragraph{Dynamics of $\lambda_S$:} 
We show that 
$$0\leq \absr{\lambda_{S,t}^2-\frac{1+\sqrt{1-4\eta}}{2} }\leq (1-\stepsize C_3)^t \absr{\delta^2 -\frac{1+\sqrt{1-4\eta}}{2} }$$
for any step size $\stepsize\leq C_4,$ where $C_3, C_4$ are two positive constants.

There are two cases to consider: when the initialization scale $\delta^2\in [1/2, \frac{1+\sqrt{1-4\eta}}{2}],$ we prove 
$$0\leq \frac{1+\sqrt{1-4\eta}}{2} -\lambda_{B,t}^2\leq (1-\stepsize C_3)^t \pr{\frac{1+\sqrt{1-4\eta}}{2} -\delta^2 };$$ 
when the initialization scale $\delta^2> \frac{1+\sqrt{1-4\eta}}{2},$ we prove 
$$0\leq  \lambda_{B,t}^2-\frac{1+\sqrt{1-4\eta}}{2} \leq (1-\stepsize C_3)^t \pr{\delta^2-\frac{1+\sqrt{1-4\eta}}{2} }.$$ 
We focus on the second case; the proof for the first case is similar. 

According to the gradient update, we have
\begin{align*}
    \lambda_{S,t+1} 
    =& \lambda_{S,t} + \stepsize \lambda_{S,t}\br{-\lambda_{S,t}^4 + \lambda_{S,t}^2 -\eta}\\
    =& \lambda_{S,t} - \stepsize \lambda_{S,t}\pr{\lambda_{S,t}^2-\frac{1-\sqrt{1-4\eta}}{2} }\pr{\lambda_{S,t}^2-\frac{1+\sqrt{1-4\eta}}{2}}
\end{align*}
We only need to show that $\lambda_{S,t}\pr{\lambda_{S,t}^2-\frac{1-\sqrt{1-4\eta}}{2} }=\Theta(1)$ for any $\lambda_{S,t}^2\in [\frac{1+\sqrt{1-4\eta}}{2},\delta].$ This is true because $\eta \in \pr{ \frac{1+\sigma^2/4}{4(1+\sigma^2)}, \frac{1+3\sigma^2/4}{4(1+\sigma^2)}}$ and $\sigma^2,\delta$ are two positive constants.

Overall, we know that there exists constant step size such that after $t=O(\log(1/\acc))$ steps, we have 
$$0\leq \lambda_{B,t}\leq \acc \text{ and } \absr{\lambda_{S,t}-\sqrt{\frac{1+\sqrt{1-4\eta}}{2}}}\leq \acc.$$
This then implies,
$$\n{W_t-\sqrt{\frac{1+\sqrt{1-4\eta}}{2}}P_S }\leq \acc.$$
\end{proofof}

\begin{proofof} {Lemma~\ref{lem:coupling}}
We know the update on $\tw_t$ is
$$\widetilde{W}_{t+1} - \twt = \stepsize \tw_{p,t}^\top \pr{-\tw_{p,t} \tw_t\pr{\frac{1}{n}\sum_{i=1}^n x_1^{(i)}[x_1^{(i)}]^\top } + \tw_{a,t} \pr{\frac{1}{n}\sum_{i=1}^n x_1^{(i)}[x_2^{(i)}]^\top }} - \stepsize \eta\tw_t,$$
and the update on $\wt$ is 
$$W_{t+1} - \wt = \stepsize W_{p,t}^\top \pr{-W_{p,t} W_t\pr{I+\sigma^2 P_B } + W_{a,t}} - \stepsize \eta W_t.$$
Next, we bound $\n{\widetilde{W}_{t+1} - \twt-\pr{W_{t+1} - \wt}}.$ According to Lemma~\ref{lem:data_matrix}, we know with probability at least $1-O(d^2) \exp\pr{-\Omega(\acc'^2 n/d^2) },$
$$\n{\frac{1}{n}\sum_{i=1}^n x_1^{(i)}[x_1^{(i)}]^\top-I-\sigma^2 P_B }, \n{\frac{1}{n}\sum_{i=1}^n x_1^{(i)}[x_2^{(i)}]^\top-I },\n{\frac{1}{n}\sum_{i=1}^n x^{(i)}[x^{(i)}]^\top-I }\leq \acc'.$$
Recall that we set $\twat=\twt$ and set $\wat$ as $\wt,$ so we have $\n{\twat-\wat} = \n{\twt-\wt}.$ Also since we set $\twpt = \twt\pr{\frac{1}{n}\sum_{i=1}^n x^{(i)}[x^{(i)}]^\top}\twt^\top$ and set $\wpt = \wt \wt^\top,$ we have $\n{\twpt-\wpt}=O\pr{\n{\twt-\wt}+\acc'}$ since $\n{W_t}=O(1).$

Combing the above bounds and recall $\stepsize$ is a constant, we have
$$\n{\widetilde{W}_{t+1} - \twt-\pr{W_{t+1} - \wt}}= O\pr{\n{\tw_t-W_t }+\acc'}.$$
Therefore, 
$$\n{\tw_t-W_t }\leq C_1^t \acc',$$
where $C_1$ is a constant larger than $1$. So for any $t\leq C\log(1/\acc),$ we have 
$$\n{\tw_t-W_t }\leq C_1^{C\log(1/\acc)} \acc' \leq (1/\acc)^{C_2} \acc',$$
for some positive constant $C_2.$ Choosing $\acc'=\acc^{C_2+1},$ we know as long as $n\geq \poly(d,1/\acc),$ with probability at least $0.99,$ for any $t\leq C\log(1/\acc),$ we have 
$$\n{\twt- \wt}\leq \acc.$$
\end{proofof}

\subsection{Sample Complexity on Down-stream Tasks}\label{sec:proofs_down_stream}
In this section, we give a proof for Theorem~\ref{thm:downstream}, which shows that the learned representations can indeed reduce sample complexity in downstream tasks.

\downstream*

Suppose $\{(z^{(i)}, y^{(i)})\}_{i=1}^n$ are $n$ training samples in the downstream task, let $Z\in \R^{n\times d}$ be the data matrix with its $i$-th row equal to $z^{(i)}.$ Denote $y \in \R^n$ as the label vector with its $i$-th entry as $y^{(i)}.$ Each input $z^{(i)}$ is transformed by a matrix $\hP\in\R^{d\times d}$ to get its representation $\hP z^{(i)}.$ The regularized loss can be written as
$$L(w) := \frac{1}{2n}\ns{Z\hP w-y } +\frac{\rho}{2}\ns{w}.$$
This is the ridge regression problem on inputs $\{(\hP z^{(i)}, y^{(i)})\}_{i=1}^n$, and the unique global minimizer $\hw$ has the following close form:
\begin{align}
    \hw = \pr{\frac{1}{n}\hP^\top Z^\top Z \hP +\rho I }^{-1}\frac{1}{n}\hP^\top Z^\top y\label{eqn:hatw}
\end{align}

With the above closed form of $\hw$, the proof of Theorem~\ref{thm:downstream}
follows by bounding the difference between $\hP \hw$ and $w^*$ by matrix concentration inequalities and matrix perturbation bounds.  Some proofs of technical lemmas are left in Appendix~\ref{sec:technical}.

\begin{proofof}{Theorem~\ref{thm:downstream}}
Denoting $\hP$ as $P+\Delta,$ we know $\fn{\Delta}\leq \acc$ by assumption. We can also write $y$ as $Zw^* + \xi$ where $\xi\in \R^n$ is the noise vector with its $i$-th entry equal to $\xi^{(i)}.$ Then, we can divide $\hw$ into two terms,
\begin{align*}
    \hw =& \pr{\frac{1}{n}\hP^\top Z^\top Z \hP +\rho I }^{-1}\frac{1}{n}\hP^\top Z^\top y\\
    =& \pr{\frac{1}{n}\hP^\top Z^\top Z \hP +\rho I }^{-1}\frac{1}{n} P^\top Z^\top \pr{Zw^*+\xi }
    +\pr{\frac{1}{n}\hP^\top Z^\top Z \hP +\rho I }^{-1}\frac{1}{n} \Delta^\top Z^\top \pr{Zw^*+\xi }
\end{align*}

Let's first give an upper bound for the second term that comes from the error term $\Delta^\top.$

\paragraph{Upper bounding $\n{\pr{\frac{1}{n}\hP^\top Z^\top Z \hP +\rho I }^{-1}\frac{1}{n} \Delta^\top Z^\top \pr{Zw^*+\xi } }$}

We first bound the norm of $\frac{1}{n} \Delta^\top Z^\top Z w^*.$ According to Lemma~\ref{lem:bound_fnorm}, we know $\whp$, $\fn{\frac{1}{\sqrt{n}} \Delta^\top Z^\top}\leq O(\acc).$ Since $Zw^*$ is a standard Gaussian vector with dimension $n$, according to Lemma~\ref{lem:norm_vector}, $\whp$, $\n{\frac{1}{\sqrt{n}}Zw^*}\leq O(1).$ Therefore, we have $\n{\frac{1}{n} \Delta^\top Z^\top Z w^*}\leq O(\acc).$

Then we bound the norm of $\frac{1}{n} \Delta^\top Z^\top \xi$. According to Lemma~\ref{lem:norm_vector}, we know $\whp,$ $\n{\frac{1}{\sqrt{n}}\xi}\leq O(\beta).$ According to Lemma~\ref{lem:bound_l2_norm}, we know with probability at least $1-\zeta/3$, $\n{\Delta^\top Z^\top \bar{\xi} }\leq O\pr{\acc\sqrt{\log(1/\zeta)}}. $ 
Therefore, we have 
$\n{\frac{1}{n} \Delta^\top Z^\top \xi}\leq O\pr{\frac{\beta \acc \sqrt{\log(1/\zeta)}}{\sqrt{n}}}. $

Since $\lambda_{\min}\pr{\frac{1}{n}\hP^\top Z^\top Z \hP +\rho I }\geq \rho,$ we have $\n{\pr{\frac{1}{n}\hP^\top Z^\top Z \hP +\rho I }^{-1} }\leq \frac{1}{\rho}.$ Combining with above bound on $\n{\frac{1}{n} \Delta^\top Z^\top \pr{Zw^*+\xi } }$, we know $\whp-\zeta/3$,
$$\n{\pr{\frac{1}{n}\hP^\top Z^\top Z \hP +\rho I }^{-1}\frac{1}{n} \Delta^\top Z^\top \pr{Zw^*+\xi } }\leq O\pr{\frac{\acc}{\rho} + \frac{\beta \acc \sqrt{\log(1/\zeta)}}{\rho \sqrt{n}}}.$$

\paragraph{Analyzing $\pr{\frac{1}{n}\hP^\top Z^\top Z \hP +\rho I }^{-1}\frac{1}{n} P^\top Z^\top \pr{Zw^*+\xi }$}

We can write $\frac{1}{n}\hP^\top Z^\top Z \hP$ as $\frac{1}{n} P^\top Z^\top Z P + E,$ where 
$$E= \frac{1}{n} \Delta^\top Z^\top Z P + \frac{1}{n} P^\top Z^\top Z \Delta + \frac{1}{n} \Delta^\top Z^\top Z \Delta.$$
Let's first bound the spectral norm of $ZP.$ Since $P$ is a projection matrix on an $r$-dimensional subspace $S,$ we can write $P$ as $UU^\top,$ where $U\in \R^{d\times r}$ has columns as an orthonormal basis of subspace $S.$ According to Lemma~\ref{lem:well_conditioned}, we know $\whp$, 
$$\Omega(1)\leq \sigma_{\min}\pr{\frac{1}{\sqrt{n}}ZU}\leq \sigma_{\max}\pr{\frac{1}{\sqrt{n}}ZU}\leq O(1).$$
Since $\n{U}\leq 1,$ we have $\n{\frac{1}{\sqrt{n}}ZP } = \n{\frac{1}{\sqrt{n}}ZUU^\top}\leq O(1).$

According to Lemma~\ref{lem:bound_fnorm}, we know $\whp,$ 
$$\fn{\frac{1}{\sqrt{n}}Z\Delta }\leq O(\acc).$$
So overall, we know $\n{E}\leq \fn{E}\leq O(\acc).$ 

Then, we can write 
$$\pr{\frac{1}{n}\hP^\top Z^\top Z \hP +\rho I }^{-1} = \pr{\frac{1}{n}P^\top Z^\top Z P +\rho I }^{-1} +F.$$
According to the perturbation bound for matrix inverse (Lemma~\ref{lem:inverse_perturb}), we have 
$\n{F}\leq O(\frac{\acc}{\rho^2}).$ Then, we have
\begin{align*}
    \pr{\frac{1}{n}\hP^\top Z^\top Z \hP +\rho I }^{-1}\frac{1}{n} P^\top Z^\top \pr{Zw^*+\xi } 
    =& \pr{\frac{1}{n} P^\top Z^\top Z P +\rho I }^{-1}\frac{1}{n} P^\top Z^\top Zw^*\\
    &+ F\frac{1}{n} P^\top Z^\top Zw^*\\
    &+ \pr{\pr{\frac{1}{n} P^\top Z^\top Z P +\rho I }^{-1}+F}\frac{1}{n} P^\top Z^\top \xi
\end{align*}

We first show that the first term is close to $w^*.$ Let the eigenvalue decomposition of $\frac{1}{n} P^\top Z^\top Z P$ be $V\Sigma V^\top,$ where $V$'s columns are an orthonormal basis for subspace $S$. Here $\Sigma\in \R^{r\times r}$ is the diagonal matrix that contains all the eigenvalues of $\frac{1}{n} P^\top Z^\top Z P$. According to Lemma~\ref{lem:well_conditioned}, we know that $\whp,$ all the non-zero eigenvalues of $\frac{1}{n} P^\top Z^\top Z P$ are $\Theta(1).$

Then, it's not hard to show that 
$$\n{\pr{\frac{1}{n}P^\top Z^\top Z P +\rho I }^{-1}\frac{1}{n}P^\top Z^\top Z P-P }\leq O(\rho).$$
This immediately implies that 
$$\n{\pr{\frac{1}{n}P^\top Z^\top Z P +\rho I }^{-1}\frac{1}{n}P^\top Z^\top Z w^*-w^* }\leq O(\rho) $$

Next, we bound the norm of the second term $F\frac{1}{n} P^\top Z^\top Zw^*.$ Similar as before, we know $\whp,$ $\n{\frac{1}{\sqrt{n}}Zw^* }\leq O(1)$ and $\n{ \frac{1}{\sqrt{n}}P^\top Z^\top}\leq O(1).$ Therefore, we have
$$\n{F\frac{1}{n} P^\top Z^\top Zw^* }\leq \n{F}\n{\frac{1}{\sqrt{n}}P^\top Z^\top }\n{ \frac{1}{\sqrt{n}}Zw^*}\leq O\pr{\frac{\acc}{\rho^2}}.$$

Finally, let's bound the third term $\pr{\pr{\frac{1}{n} P^\top Z^\top Z P +\rho I }^{-1}+F}\frac{1}{n} P^\top Z^\top \xi.$ We first bound the norm of $\frac{1}{n}P^\top Z^\top \xi.$ $\whp,$ we know $\n{\xi}\leq 2\beta\sqrt{n}.$ Therefore, we know $\n{\frac{1}{n}P^\top Z^\top \xi}\leq O(\beta/\sqrt{n})\n{P^\top Z^\top \bar{\xi} },$ where $\bar{\xi} = \xi/\n{\xi}.$ According to Lemma~\ref{lem:P_noise}, with probability at least $1-\zeta/3,$ we have $\n{P^\top Z^\top \bar{\xi} }\leq \sqrt{r}+O(\sqrt{\log(1/\zeta)}).$
Overall, $\whp-\zeta/3,$
$$\n{\frac{1}{n} P^\top Z^\top \xi}\leq O\pr{\frac{\sqrt{r}\beta  + \sqrt{\log(1/\zeta)} \beta }{\sqrt{n}}}.$$
It's not hard to verify that for any vector $v\in \R^d$ in the subspace $S,$ we have $\n{\pr{\pr{\frac{1}{n} P^\top Z^\top Z P +\rho I }^{-1}+F}v}\leq O(\n{v}).$ Since $\frac{1}{n} P^\top Z^\top \xi$ lies on subspace $S,$ we have 
$$\n{\pr{\pr{\frac{1}{n} P^\top Z^\top Z P +\rho I }^{-1}+F}\frac{1}{n} P^\top Z^\top \xi }\leq O\pr{\frac{\sqrt{r}\beta  + \sqrt{\log(1/\zeta)}\beta }{\sqrt{n}}}.$$

Combining the above analysis and taking a union bound over all the events, we know $\whp-2\zeta/3,$
\begin{align*}
    \n{\hw-w^*} = O\pr{\rho + \frac{\acc}{\rho} + \frac{\acc}{\rho^2} +\frac{\beta \acc \sqrt{\log(1/\zeta)}}{\rho \sqrt{n}} + \frac{\sqrt{r}\beta + \sqrt{\log(1/\zeta)}\beta }{\sqrt{n}}}
\end{align*}
Suppose $n\geq O(\log(1/\zeta))$ and setting $\rho = \acc^{1/3},$ we further have with probability at least $1-\zeta,$
\begin{align*}
    \n{\hw-w^*} =& O\pr{\acc^{1/3} +\frac{\beta \acc^{2/3} \sqrt{\log(1/\zeta)}}{ \sqrt{n}} + \frac{\sqrt{r}\beta + \sqrt{\log(1/\zeta)}\beta }{\sqrt{n}}}\\
    \leq & O\pr{\acc^{1/3} + \beta\frac{\sqrt{r} + \sqrt{\log(1/\zeta)} }{\sqrt{n}}},
\end{align*}
where the last inequality assumes $\acc<1.$ 

We can also bound $\n{\hP \hw - w^*}$ as follows,
\begin{align*}
    \n{\hP \hw - w^*} = &\n{\hP \hw - P\hw + P\hw -P w^*}\\
    \leq &\n{\hP \hw - P\hw} + \n{P\hw -P w^*}\\
    \leq &\n{\hP-P}\n{\hw} + \n{P}\n{\hw-w^*}\\
    \leq & \acc O\pr{1+\acc^{1/3} + \beta\frac{\sqrt{r} + \sqrt{\log(1/\zeta)} }{\sqrt{n}}} + O\pr{\acc^{1/3} + \beta\frac{\sqrt{r} + \sqrt{\log(1/\zeta)} }{\sqrt{n}}}\\
    \leq & O\pr{\acc^{1/3} + \beta\frac{\sqrt{r} + \sqrt{\log(1/\zeta)} }{\sqrt{n}}}
\end{align*}
\end{proofof}

\subsection{Analysis with $W_p:=(W\E_{x_1}x_1x_1^\top W^\top)^\alpha$}\label{sec:diff_corr}

In this section, we prove that DirectSet($\alpha$) can also learn the projection matrix when we set $W_p:=(W\E_{x_1}x_1x_1^\top W^\top)^\alpha$. For the network architecture and data distribution, we follow exactly the same setting as in Section~\ref{sec:GF_population}. Therefore, we know $W_p:=(W\E_{x_1}x_1x_1^\top W^\top)^\alpha = (W(I+\sigma^2 P_B) W^\top)^\alpha$.

\begin{theorem}\label{thm:diff_corr}
Suppose network architecture and data distribution are as defined in Assumption~\ref{assump:network} and Assumption~\ref{assump:data_distribution}, respectively. Suppose we initialize online network $W$ as $\delta I,$ and run DirectPred$(\alpha)$ on population loss (see Eqn.~\ref{eqn:population_loss}) with infinitesimal step size and $\eta$
weight decay. Suppose we set $W_a =W$ and $W_p=(W\E_{x_1}x_1x_1^\top W^\top)^\alpha.$ Assuming the weight decay coefficient $\eta \in \pr{ \frac{1}{4(1+\sigma^2)^{1+2\alpha}}, \frac{1}{4}}$ and initialization scale $\delta>\pr{\frac{1-\sqrt{1-4\eta}}{2}}^{1/(2\alpha)},$ we know $W$ converges to $\pr{\frac{1+\sqrt{1-4\eta}}{2}}^{1/(2\alpha)}P_S$ when time goes to infinity.
\end{theorem}

The only difference from Theorem~\ref{thm:diff_corr} is that now the initialization $\delta$ is only required to be larger than $\frac{1}{4(1+\sigma^2)^{1+2\alpha}}$. The proof is almost the same as in Theorem~\ref{thm:learn_subspace}.
 
\begin{proofof}{Theorem~\ref{thm:diff_corr}}
Similar as in the proof of Theorem~\ref{thm:learn_subspace}, we can write the dynamics on $W$ is as follows:
\begin{align*}
    \dot W = & W_p^\top (-W_p W(I+\sigma^2 P_B) + W_a) - \eta W\\
     = & \absr{W^2(I+\sigma^2 P_B)}^{\alpha} (-\absr{W^2(I+\sigma^2 P_B)}^{\alpha} W(I+\sigma^2 P_B) + W) - \eta W\\
    =& W\pr{-(I+\sigma^2 P_B)^{1+2\alpha} \absr{W}^{4\alpha} + \absr{W}^{2\alpha} -\eta}.
\end{align*}

\paragraph{Dynamics for $\lambda_B$:} We can write down the dynamics for $\lambda_B$ as follows:
$$
\dot \lambda_B = \lambda_B\br{-(1+\sigma^2)^{1+2\alpha}\absr{\lambda_B}^{4\alpha} + \absr{\lambda_B}^{2\alpha} -\eta}
$$
When $\eta>\frac{1}{4(1+\sigma^2)^{1+2\alpha}},$ we know $\dot \lambda_B <0$ for any $\lambda_B >0$ and $\lambda_B = 0$ is a critical point. This means, as long as $\eta>\frac{1}{4(1+\sigma^2)^{1+2\alpha}}$, $\lambda_B$ must converge to zero.

\paragraph{Dynamics for $\lambda_S$:} The dynamics is same as when setting $W_p=(WW^\top)^\alpha,$
$$
\dot \lambda_S = \lambda_S\br{-\absr{\lambda_S}^{4\alpha} + \absr{\lambda_S}^{2\alpha} -\eta}.
$$
so when $0<\eta<\frac{1}{4}$ and initialization $\delta^{2\alpha} > \frac{1-\sqrt{1-4\eta}}{2}$, we know $\lambda_S^{2\alpha}$ converges to $\frac{1+\sqrt{1-4\eta}}{2}.$

Overall, we know when $\frac{1}{4(1+\sigma^2)^{1+2\alpha}}<\eta<\frac{1}{4}$ and $\delta>\pr{\frac{1-\sqrt{1-4\eta}}{2}}^{1/(2\alpha)},$ we have $\lambda_B$ converge to zero and $\lambda_S$ converge to $\pr{\frac{1+\sqrt{1-4\eta}}{2}}^{1/(2\alpha)}.$ That is, matrix $W$ converges to $\pr{\frac{1+\sqrt{1-4\eta}}{2}}^{1/(2\alpha)}P_S.$
\end{proofof}

\subsection{Technical Lemmas}\label{sec:technical}
\begin{lemma}\label{lem:data_matrix}
Suppose $\{x^{(i)},x_1^{(i)},x_2^{(i)}\}_{i=1}^n$ are sampled as decribed in Section~\ref{sec:two_layer}.
Suppose $n\geq O(d/\acc^2),$ with probability at least $1-O(d^2) \exp\pr{-\Omega(\acc^2 n/d^2) },$
we have 
$$\n{\frac{1}{n}\sum_{i=1}^n x_1^{(i)}[x_1^{(i)}]^\top-I-\sigma^2 P_B }, \n{\frac{1}{n}\sum_{i=1}^n x_1^{(i)}[x_2^{(i)}]^\top-I }, \n{\frac{1}{n}\sum_{i=1}^n x^{(i)}[x^{(i)}]^\top-I }\leq \acc.$$
\end{lemma}
\begin{proofof}{Lemma~\ref{lem:data_matrix}}
For each $x_1^{(i)}$, we can write it as $x^{(i)}+z_1^{(i)}$ where $x^{(i)}\sim\mathcal{N}(0,I)$ and $z_1^{(i)}\sim \mathcal{N}(0,\sigma^2 P_B).$ So we have 
$$\frac{1}{n}\sum_{i=1}^n x_1^{(i)}[x_1^{(i)}]^\top = \frac{1}{n}\sum_{i=1}^n \pr{x^{(i)}[x^{(i)}]^\top + z_1^{(i)}[z_1^{(i)}]^\top + x^{(i)}[z_1^{(i)}]^\top+z_1^{(i)}[x^{(i)}]^\top }.$$

According to Lemma~\ref{lem:sig_matrix}, we know as long as $n\geq O(d/\acc^2),$ with probability at least $1-\exp(-\Omega(\acc^2 n)),$
$$\n{\frac{1}{n}\sum_{i=1}^n x^{(i)}[x^{(i)}]^\top - I}\leq \acc.$$
Similarly, with probability at least $1-\exp(-\Omega(\acc^2 n)),$
$$\n{\frac{1}{n}\sum_{i=1}^n z_1^{(i)}[z_1^{(i)}]^\top - \sigma^2 P_B}\leq \acc.$$ 

Next we bound $\n{\frac{1}{n}\sum_{i=1}^n x^{(i)}[z_1^{(i)}]^\top}.$ We know each entry in matrix $\frac{1}{n}\sum_{i=1}^n x^{(i)}[z_1^{(i)}]^\top$ is the average of $n$ zero-mean $O(1)$-subexponential independent random variables. Therefore, according to the Bernstein's inequality, for any fixed entry $(k,l),$ with probability at least $1-\exp\pr{-\acc^2 n/d^2 },$
$$\absr{\br{\frac{1}{n}\sum_{i=1}^n x^{(i)}[z_1^{(i)}]^\top}_{k,l} }\leq \acc/d.$$
Taking a union bound over all the entries, we know with probability at least $1-d^2 \exp\pr{-\acc^2 n/d^2 },$ 
$$\n{\frac{1}{n}\sum_{i=1}^n x^{(i)}[z_1^{(i)}]^\top}\leq \fn{\frac{1}{n}\sum_{i=1}^n x^{(i)}[z_1^{(i)}]^\top}\leq \acc.$$
The same analysis also applies to $\n{\frac{1}{n}\sum_{i=1}^n z_1^{(i)}[x^{(i)}]^\top }.$
Combing all the bounds, we know with probability at least $1-O(d^2) \exp\pr{-\Omega(\acc^2 n/d^2) },$
$$\n{\frac{1}{n}\sum_{i=1}^n x_1^{(i)}[x_1^{(i)}]^\top-I-\sigma^2 P_B }\leq 4\acc.$$

Similarly, we can prove that with probability at least $1-O(d^2) \exp\pr{-\Omega(\acc^2 n/d^2) },$
$$\n{\frac{1}{n}\sum_{i=1}^n x_1^{(i)}[x_2^{(i)}]^\top-I}\leq 4\acc.$$
Changing $\acc$ to $\acc'/4$ finishes the proof.
\end{proofof}

\begin{lemma}\label{lem:well_conditioned}
Let $X\in \R^{n\times d}$ be a standard Gaussian matrix, and let $U\in \R^{d\times r}$ be a matrix with orthonormal columns. Suppose $n\geq 2r$, with probability at least $1-\exp(-\Omega(n))$, we know 
$$\Omega(1)\leq \lambda_{\min}\pr{\frac{1}{n}U^\top X^\top X U }\leq \lambda_{\max}\pr{\frac{1}{n}U^\top X^\top X U }\leq O(1). $$
\end{lemma}
\begin{proofof}{Lemma~\ref{lem:well_conditioned}}
Since $U$ has orthonormal columns, we know $XU$ is a $n\times r$ matrix with each entry independently sampled from $\mathcal{N}(0,1).$ According to Lemma~\ref{lem:sig_matrix}, we know when $n\geq 2r,$ with probability at least $1-\exp(-\Omega(n)),$ 
$$ \Omega(1)\leq \sigma_{\min}\pr{\frac{1}{\sqrt{n}} X U }\leq \sigma_{\max}\pr{\frac{1}{\sqrt{n}} X U } \leq O(1).$$
This immediately implies that
$$\Omega(1)\leq \lambda_{\min}\pr{\frac{1}{n}U^\top X^\top X U }\leq \lambda_{\max}\pr{\frac{1}{n}U^\top X^\top X U }\leq O(1).$$
\end{proofof}

\begin{lemma}\label{lem:bound_fnorm}
Let $\Delta$ be a $d\times d$ matrix with Frobenius norm $\acc$, and let $X$ be a $n\times d$ standard Gaussian matrix. We know $\whp$,
$$\fn{\frac{1}{\sqrt{n}} X\Delta}\leq O(\acc).$$
\end{lemma}
\begin{proofof}{Lemma~\ref{lem:bound_fnorm}}
Let the singular value decomposition of $\Delta$ be $U\Sigma V^\top,$ where $U, V$ have orthonormal columns and $\Sigma$ is a diagonal matrix with diagonals equal to singular values $\sigma_i$'s. Since $\fn{\Delta}=\acc,$ we know $\sum_{i=1}^d \sigma_i^2 = \acc^2.$

Since $U$ is an orthonormal matrix, we know $\hat{X}:=XU$ is still an $n\times d$ standard Gaussian matrix. Next, we bound the Frobenius norm of $\widetilde{X}:= \hat{X}\Sigma.$ It's not hard to verify that all the entries in $\tX$ are independent Gaussian variables and $\tX_{ij}\sim \mathcal{N}(0, \sigma_j^2).$ According to the Bernstein's inequality for sum of independent and sub-exponential random variables, we have for every $t>0,$
$$\Pr\br{\absr{\sum_{i\in [n],j\in [d]}\tX_{ij}^2 - n\acc^2}\geq t }\leq 2\exp\br{-c\min\pr{\frac{t^2}{\sum_{i\in [n],j\in [d]}\sigma_j^4}, \frac{t}{\max_{j\in [d]} \sigma_j^2} }}.$$
Since $\sum_{j=1}^d \sigma_j^2 = \fns{\Delta} = \acc^2,$ we know $\max_{j\in [d]}\sigma_j^2 \leq \acc^2.$ We also have $\sum_{j\in [d]}\sigma_j^4 \leq \pr{\sum_{j\in [d]}\sigma_j^2 }^2= \acc^4.$
Therefore, we have 
$$\Pr\br{\absr{\sum_{i\in [n],j\in [d]}\tX_{ij}^2 - n\acc^2}\geq t }\leq 2\exp\br{-c\min\pr{\frac{t^2}{n\acc^4}, \frac{t}{\acc^2} }}.$$
Replacing $t$ by $n\acc^2,$ we concluded that $\whp$,$\fns{\tX}\leq 2n\acc^2.$ Furthermore, since $\n{V^\top}  =1, $ we have 
$$\fn{\frac{1}{\sqrt{n}}X\Delta} = \fn{\frac{1}{\sqrt{n}}\tX V^\top}\leq \fn{\frac{1}{\sqrt{n}}\tX}\n{V}\leq O(\acc).$$
\end{proofof}

\begin{lemma}\label{lem:bound_l2_norm}
Let $\Delta^\top$ be a $d\times d$ matrix with Frebenius norm $\acc$ and let $X^\top$ be a $d\times n$ standard Gaussian matrix. Let $\bar{\xi}$ be a unit vector with dimension $n$. We know with probability at least $1-\zeta/3,$
$$\n{\Delta^\top X^\top \bar{\xi} }\leq O(\acc\sqrt{\log(1/\zeta})). $$
\end{lemma}
\begin{proofof}{Lemma~\ref{lem:bound_l2_norm}}
Let the sigular value decomposition of $\Delta^\top$ be $U\Sigma V^\top.$ We know $X^\top \bar{\xi}$ is a $d$-dimensional standard Gaussian vector. Further, we know $V^\top X^\top \bar{\xi}$ is also a $d$-dimensional standard Gaussian vector. So $\Sigma V^\top X^\top \bar{\xi}$ has independent Gaussian entries with its $i$-th entry distributed as $\mathcal{N}(0, \sigma_i^2).$
According to the Bernstein's inequality for sum of independent and sub-exponential random variables, we have for every $t>0,$
$$\Pr\br{\absr{\ns{\Sigma V^\top X^\top \bar{\xi}} -\acc^2}\geq t }\leq 2\exp\br{-c\min\pr{\frac{t^2}{\acc^4}, \frac{t}{\acc^2} }}.$$
Choosing $t$ as $O(\acc^2\log(1/\zeta)),$ we know with probability at least $1-\zeta/3,$ we have 
$$\ns{\Sigma V^\top X^\top \bar{\xi}}\leq O\pr{\acc^2 \log(1/\zeta)} .$$
Since $\n{U}=1,$ we further have
$$\n{\Delta^\top X^\top \bar{\xi} } = \n{ U\Sigma V^\top X^\top \bar{\xi}}\leq  \n{ U}\n{\Sigma V^\top X^\top \bar{\xi}}\leq O\pr{\acc\sqrt{\log(1/\zeta)}}$$
\end{proofof}

\begin{lemma}\label{lem:P_noise}
Let $P\in \R^{d\times d}$ be a projection matrix on a $r$-dimensional subspace, and let $\bar{\xi}$ be a unit vector in $\R^d.$ Let $X^\top$ be a $d\times n$ standard Gaussian matrix that is independent with $P$ and $\xi$. With probability at least $1-\zeta/3,$ we have 
$$\n{P^\top X^\top \bar{\xi} }\leq \sqrt{r}+O(\sqrt{\log(1/\zeta)}).$$
\end{lemma}
\begin{proofof}{Lemma~\ref{lem:P_noise}}
Since $P$ is a projection matrix on an $r$-dimensional subspace, we can write $P$ as $UU^\top,$ where $U\in \R^{d\times r}$ has orthonormal columns. We know $U^\top X^\top$ is still a standard Gaussian matrix with dimension $r\times n.$ Furthermore, $U^\top X^\top \bar{\xi}$ is an $r$-dimensional standard Gaussian vector. According to Lemma~\ref{lem:norm_vector}, with probability at least $1-\zeta/3,$ we have 
$$\n{U^\top X^\top \bar{\xi}}\leq \sqrt{r} + O(\sqrt{\log(1/\zeta)}).$$
Since $\n{U}=1,$ we further have 
$$\n{P^\top X^\top \bar{\xi}} = \n{UU^\top X^\top \bar{\xi}} \leq \n{U}\n{U^\top X^\top \bar{\xi}}\leq \sqrt{r} + O(\sqrt{\log(1/\zeta)}).$$
\end{proofof}
\section{Analysis of Deep Linear Networks}\label{sec:deep_linear}
In this section, we extend the analysis in Section~\ref{sec:GF_population} to deep linear networks. We consider the same data distribution as defined in Assumption~\ref{assump:data_distribution}.
We consider the following network,
\begin{assumption}[Deep linear network]\label{assump:deep_linear}
The online network is an $l$-layer linear networks $W_l W_{l-1}\cdots W_1$ with each $W_i\in \R^{d\times d}.$ The target network has the same architecture with weight matrices $W_{a,l}W_{a,l-1}\cdots W_{a,1}.$ For convenience, we denote $W$ as $W_l W_{l-1}\cdots W_1$ and denote $W_a$ as $W_{a,l}W_{a,l-1}\cdots W_{a,1}.$
\end{assumption}

 \paragraph{Training procedure: }At the initialization, we initialize each $W_i$ as $\delta^{1/l}I_d$. Through the training, we fix $W_p$ as $\pr{WW^\top}^\alpha$ and fix each $W_{a,i}$ as $W_i.$ We run gradient flow on every $W_i$ with weight decay $\eta.$ The population loss is
$$L(\{W_i\},W_p,\{W_{a,i}\}) := \frac{1}{2}\E_{x_1,x_2} \ns{W_p W_l W_{l-1}\cdots W_1x_1 - \stopgrad(W_{a,l}W_{a,l-1}\cdots W_{a,1} x_2)}.$$

\begin{theorem}\label{thm:deep_linear}
Suppose the data distribution and network architecture satisfies Assumption~\ref{assump:data_distribution} and Assumption~\ref{assump:deep_linear}, respectively. Suppose we train the network as described above. Assuming the weight decay coefficient \\$\eta\in \pr{\frac{2\alpha l(2\alpha l+2l-2)^{1+\frac{1}{\alpha}-\frac{1}{\alpha l}}}{(4\alpha l+2l-2)^{2+\frac{1}{\alpha}-\frac{1}{\alpha l}}(1+\sigma^2)^{1+\frac{1}{\alpha}-\frac{1}{\alpha l}}}, \frac{2\alpha l(2\alpha l+2l-2)^{1+\frac{1}{\alpha}-\frac{1}{\alpha l}}}{(4\alpha l+2l-2)^{2+\frac{1}{\alpha}-\frac{1}{\alpha l}}}},$ and initialization scale $\delta\geq \pr{\frac{2\alpha l+2l-2}{4\alpha l+2l-2}}^{\frac{1}{2\alpha}},$ we know $W$ converges to $cP_S$ as time goes to infinity, where $c$ is a positive number within $\pr{\pr{\frac{2\alpha l+2l-2}{4\alpha l+2l-2}}^{\frac{1}{2\alpha}}, 1}$.
\end{theorem}

Similar as in the setting of single-layer linear networks, we prove Theorem~\ref{thm:deep_linear} by analyzing the dynamics of the eigenvalues of $W.$ Note that with constant $\alpha$, the upper/lower bounds for $\eta$ and scalar $c$ in the Theorem are always constants no matter how large $l$ is.

\begin{proofof}{Theorem~\ref{thm:deep_linear}}
For $j\geq i$, we use $W_{[j:i]}$ to denote $W_j W_{j-1}\cdots W_i$ and for $j<i$ have $W_{[j:i]} = I.$ We use similar notations for $W_{a,[j:i]}.$
For each $W_i,$ we can compute its dynamics as follows:
\begin{align*}
    \dot W_i = &-\pr{W_p W_{[l:i+1]}}^\top \pr{W_p W(I+\sigma^2 P_B)} \pr{W_{[i-1:1]}}^\top + \pr{W_p W_{a,[l:i+1]}}^\top W_a \pr{W_{a,[i-1:1]}}^\top - \eta W_i.
\end{align*}
It's clear that through the training all $W_i$'s remains the same and they are simultaneously diagonalizable with $W_p, I$ and $P_B$. We also have $W_a = W$ and $W_p = \absr{W}^{2\alpha}.$ Since we will ensure that $W$ is always positive semi-definite so $W_p = \absr{W}^{2\alpha} = W^{2\alpha} = W_i^{2\alpha l}.$  So the dynamics for each $W_i$ can be simplified as follows:
\begin{align*}
    \dot W_i = -W_i^{4\alpha l + 2l-1}(I+\sigma^2P_B) + W_i^{2\alpha l+ 2l-1} -\eta W_i.
\end{align*}

Let the eigenvalue decomposition of $W_i$ be $\sum_{i=1}^d \nu_i u_i u_i^\top,$ with span$(\{u_{d-r+1}, \cdots, u_d\})$ equals to subspace $B$. We can separately analyze the dynamics of each $\nu_i.$ Furthermore, we know $\nu_1, \cdots, \nu_r$ have the same value $\nu_S$ and $\nu_{d-r+1},\cdots, \nu_d$ have the same value $\nu_B.$ We can write down the dynamics for $\nu_S$ and $\nu_B$ as follows,
\begin{align*}
    \dot \nu_S =& -\nu_S^{4\alpha l+2l-1} + \nu_S^{2\alpha l+2l-1} -\eta \nu_S,\\
    \dot \nu_B =& -\nu_B^{4\alpha l+2l-1}(1+\sigma^2) + \nu_B^{2\alpha l+2l-1} -\eta \nu_B.
\end{align*}
Let $\lambda_S$ be the eigenvalue of $W$ corresponding to eigen-directions $u_1,\cdots,u_r,$ and let $\lambda_B$ be the eigenvalue of $W$ corresponding to eigen-directions $u_{d-r+1},\cdots,u_d.$ We know $\lambda_S = \nu_S^l$ and $\lambda_B = \nu_B^l.$ So we can write down the dynamics for $\lambda_B$ as follows,
\begin{align*}
    \dot \lambda_B = l\nu_B^{l-1}\dot \nu_B =& -l\nu_B^{4\alpha l+3l-2}(1+\sigma^2) +l \nu_B^{2\alpha l+3l-2} -l\eta \nu_B^{l}\\
    =& -l\lambda_B^{4\alpha +3-2/l}(1+\sigma^2) +l \lambda_B^{2\alpha +3-2/l} -l\eta \lambda_B,
\end{align*}
and similarly for $\lambda_S$ we have
$$\dot \lambda_S = -l\lambda_S^{4\alpha +3-2/l} +l \lambda_S^{2\alpha +3-2/l} -l\eta \lambda_S.$$

\paragraph{Dynamics for $\lambda_B$:} We can write the dynamics on $\lambda_B$ as follows,
\begin{align*}
    \dot \lambda_B = l\lambda_B g(\lambda_B),
\end{align*}
where $g(\lambda_B):=-\lambda_B^{4\alpha +2-2/l}(1+\sigma^2) + \lambda_B^{2\alpha +2-2/l} - \eta.$ We show that when $\eta$ is large enough, $g(\lambda_B)$ is negative for any positive $\lambda_B.$ We compute the maximum value of $g(\lambda_B)$ for $\lambda_B>0.$ We first compute the derivative of $g$ as follows:
\begin{align*}
    g'(\lambda_B) =& -(4\alpha +2-2/l)(1+\sigma^2)\lambda_B^{4\alpha +1-2/l} + (2\alpha +2-2/l)\lambda_B^{2\alpha +1-2/l}\\
    =& \lambda_B^{2\alpha +1-2/l}\pr{-(4\alpha +2-2/l)(1+\sigma^2)\lambda_B^{2\alpha} + (2\alpha +2-2/l) }.
\end{align*}
It's clear that $g'(\lambda_B)>0$ for $\lambda_B^{2\alpha}\in (0, \frac{2\alpha l+2l-2}{(4\alpha l+2l-2)(1+\sigma^2)})$ and $g'(\lambda_B)<0$ for $\lambda_B^{2\alpha}\in ( \frac{2\alpha l+2l-2}{(4\alpha l+2l-2)(1+\sigma^2)},+\infty).$ Therefore, the maximum value of $g(\lambda_B)$ for positive $\lambda_B$ takes at $\lambda_B^* =  \pr{\frac{2\alpha l+2l-2}{(4\alpha l+2l-2)(1+\sigma^2)}}^{\frac{1}{2\alpha}}$ and 
\begin{align*}
    g(\lambda_B^*) =& -\pr{\frac{2\alpha l+2l-2}{(4\alpha l+2l-2)(1+\sigma^2)}}^{2+\frac{1}{\alpha}-\frac{1}{\alpha l}}(1+\sigma^2) + \pr{\frac{2\alpha l+2l-2}{(4\alpha l+2l-2)(1+\sigma^2)}}^{1+\frac{1}{\alpha}-\frac{1}{\alpha l}} - \eta\\
    =& \frac{2\alpha l(2\alpha l+2l-2)^{1+\frac{1}{\alpha}-\frac{1}{\alpha l}}}{(4\alpha l+2l-2)^{2+\frac{1}{\alpha}-\frac{1}{\alpha l}}(1+\sigma^2)^{1+\frac{1}{\alpha}-\frac{1}{\alpha l}}}-\eta.
\end{align*}
As long as $\eta> \frac{2\alpha l(2\alpha l+2l-2)^{1+\frac{1}{\alpha}-\frac{1}{\alpha l}}}{(4\alpha l+2l-2)^{2+\frac{1}{\alpha}-\frac{1}{\alpha l}}(1+\sigma^2)^{1+\frac{1}{\alpha}-\frac{1}{\alpha l}}},$ we know $g(\lambda_B)<0$ for any $\lambda_B>0$, which further implies that $\dot \lambda_B < 0$ for any $\lambda_B>0$. So $\lambda_B$ converges to zero.

\paragraph{Dynamics for $\lambda_S:$} We can write down the dynamics on $\lambda_S$ as follows, 
$$\dot \lambda_S = l\lambda_S h(\lambda_S),$$
where $h(\lambda_S) = -\lambda_S^{4\alpha+2-2/l} + \lambda_S^{2\alpha+2-2/l} - \eta.$ We compute the derivative of $h$ as follows:
$$h'(\lambda_S) = \lambda_S^{2\alpha+1-2/l}\pr{-(4\alpha+2-2/l)\lambda_S^{2\alpha} + (2\alpha+2-2/l) }.$$
So $h(\lambda_S)$ is increasing in $(0,\pr{\frac{2\alpha l+2l-2}{4\alpha l+2l-2}}^{\frac{1}{2\alpha}})$ and is decreasing in $(\pr{\frac{2\alpha l+2l-2}{4\alpha l+2l-2}}^{\frac{1}{2\alpha}}, \infty).$ The maximum value of $h$ for positive $\lambda_S$ takes at $\lambda_S^* =\pr{\frac{2\alpha l+2l-2}{4\alpha l+2l-2}}^{\frac{1}{2\alpha}}$ and we have 
$$h(\lambda_S^*) = \frac{2\alpha l(2\alpha l+2l-2)^{1+\frac{1}{\alpha}-\frac{1}{\alpha l}}}{(4\alpha l+2l-2)^{2+\frac{1}{\alpha}-\frac{1}{\alpha l}}}-\eta.$$
As long as $\eta<\frac{2\alpha l(2\alpha l+2l-2)^{1+\frac{1}{\alpha}-\frac{1}{\alpha l}}}{(4\alpha l+2l-2)^{2+\frac{1}{\alpha}-\frac{1}{\alpha l}}},$ we have $h(\lambda_S^*)>0.$ Furthermore, since $h$ is increasing in $(0,\lambda_S^*)$ and is decreasing in $(\lambda_S^*,\infty)$ and $h(0), h(\infty)<0,$ we know there exists $\lambda_S^-\in (0,\lambda_S^*),\lambda_S^+\in(\lambda_S^*,\infty)$ such that $h(\lambda_S)<0$ in $(0,\lambda_S^-)$, $h(\lambda_S)>0$ in $(\lambda_S^-,\lambda_S^+)$ and $h(\lambda_S)<0$ in $(\lambda_S^+,\infty).$ Therefore, as long as $\delta\geq \lambda_S^*>\lambda_S^-,$ we have $\lambda_S$ converges to $\lambda_S^+.$ Since $h(1)<0,$ we know $\lambda_S^+\in (\pr{\frac{2\alpha l+2l-2}{4\alpha l+2l-2}}^{\frac{1}{2\alpha}}, 1).$

Overall as long as $\eta\in \pr{\frac{2\alpha l(2\alpha l+2l-2)^{1+\frac{1}{\alpha}-\frac{1}{\alpha l}}}{(4\alpha l+2l-2)^{2+\frac{1}{\alpha}-\frac{1}{\alpha l}}(1+\sigma^2)^{1+\frac{1}{\alpha}-\frac{1}{\alpha l}}}, \frac{2\alpha l(2\alpha l+2l-2)^{1+\frac{1}{\alpha}-\frac{1}{\alpha l}}}{(4\alpha l+2l-2)^{2+\frac{1}{\alpha}-\frac{1}{\alpha l}}}},$ we know $W$ converges to $cP_S,$ where $c$ is a positive number within $(\pr{\frac{2\alpha l+2l-2}{4\alpha l+2l-2}}^{\frac{1}{2\alpha}}, 1)$.
\end{proofof}
\section{Analysis of Predictor Regularization.}\label{sec:proofs_eps_identity}
In this section, we study the influence of predictor regularization in a simple linear setting. In particular, we consider the same setting as in Section~\ref{sec:GF_population} except that we set $W_p:=(WW^\top)^\alpha+\epsilon I.$

\begin{theorem}\label{thm:eps_identity}
In the setting of Theorem~\ref{thm:learn_subspace} except that we set $W_p = (WW^\top)^\alpha+\epsilon I.$ We have
\begin{itemize}
    \item when $\epsilon\in [0,\frac{1+\sqrt{1-4\eta}}{2}),$ as long as $\delta>\pr{\max\pr{\frac{1-\sqrt{1-4\eta}}{2}-\epsilon, 0}}^{\frac{1}{2\alpha}},$ we have $W$ converges to $\pr{\frac{1+\sqrt{1-4\eta}}{2}-\epsilon}^{\frac{1}{2\alpha}}P_S$;
    \item when $\epsilon\geq \frac{1+\sqrt{1-4\eta}}{2},$ $W$ always converges to zero.
\end{itemize}
\end{theorem}

\begin{proofof}{Theorem~\ref{thm:eps_identity}}
We can write the dynamics of $W$ as follows,
\begin{align*}
\dot{W} =&  W_p^\top (-W_p W(I+\sigma^2 P_B) + W_a) - \eta W\\
    =& W\pr{-(I+\sigma^2 P_B) \pr{\absr{W}^{2\alpha}+\epsilon I}^2 + \pr{\absr{W}^{2\alpha}+\epsilon I} -\eta}.
\end{align*}

Let the eigenvalue decomposition of $W$ be $\sum_{i=1}^d \lambda_i u_i u_i^\top,$ with span$(\{u_{d-r+1}, \cdots, u_d\})$ equals to subspace $B$. We can separately analyze the dynamics of each $\lambda_i.$ Furthermore, we know $\lambda_1, \cdots, \lambda_r$ have the same value $\lambda_S$ and $\lambda_{d-r+1},\cdots, \lambda_d$ have the same value $\lambda_B.$

\paragraph{Dynamics for $\lambda_B$:} We can write down the dynamics for $\lambda_B$ as follows:
$$
\dot \lambda_B = \lambda_B\br{-(1+\sigma^2) \pr{\absr{\lambda_B}^{2\alpha}+\epsilon}^2 + \pr{\absr{\lambda_B}^{2\alpha}+\epsilon} -\eta}
$$
When $\eta>\frac{1}{4(1+\sigma^2)},$ we still know $\dot \lambda_B <0$ for any $\lambda_B >0$ and $\lambda_B = 0$ is a critical point. So $\lambda_B$ converges to zero.

\paragraph{Dynamics for $\lambda_S$:} We can write down the dynamics for $\lambda_S$ as follows:
\begin{align*}
    \dot \lambda_S =&  \lambda_S\br{-\pr{\absr{\lambda_S}^{2\alpha}+\epsilon}^2 + \pr{\absr{\lambda_S}^{2\alpha}+\epsilon} -\eta}\\
    =& -\lambda_S\pr{\absr{\lambda_S}^{2\alpha}+\epsilon-\frac{1-\sqrt{1-4\eta}}{2}}\pr{\absr{\lambda_S}^{2\alpha}+\epsilon-\frac{1+\sqrt{1-4\eta}}{2}},
\end{align*}
where the second inequality assumes $0<\eta<\frac{1}{4}.$ We have
\begin{itemize}
    \item when $\epsilon\in [0,\frac{1+\sqrt{1-4\eta}}{2}),$ as long as $\delta>\pr{\max\pr{\frac{1-\sqrt{1-4\eta}}{2}-\epsilon, 0}}^{\frac{1}{2\alpha}},$ we have $\lambda_S$ converges to $\pr{\frac{1+\sqrt{1-4\eta}}{2}-\epsilon}^{\frac{1}{2\alpha}}>0$;
    \item when $\epsilon\geq \frac{1+\sqrt{1-4\eta}}{2},$ $\lambda_S$ always converges to zero.
\end{itemize}
\end{proofof}
\section{Technical Tools}
\subsection{Norm of Random Vectors}
The following lemma shows that a standard Gaussian vector with dimension $n$ has $\ell_2$ norm concentrated at $\sqrt{n}$.

\begin{lemma}[Theorem 3.1.1 in~\cite{vershynin2018high}]\label{lem:norm_vector}
Let $X=(X_1, X_2, \cdots, X_n)\in \R^n$ be a random vector with each entry independently sampled from $\mathcal{N}(0,1).$
Then
$$\Pr[\absr{\n{x}-\sqrt{n}}\geq t]\leq 2\exp(-t^2/C^2),$$
where $C$ is an absolute constant.
\end{lemma}

\subsection{Singular Values of Gaussian Matrices}
The following lemma shows a tall random Gaussian matrix is well-conditioned with high probability.

\begin{lemma}[Corollary 5.35 in~\cite{vershynin2010introduction}]\label{lem:sig_matrix}
Let $A$ be an $N\times n$ matrix whose entries are independent standard normal random variables. Then for every $t\geq 0$ with probability at least $1-2\exp(-t^2/2)$ one has 
$$\sqrt{N}-\sqrt{n}-t \leq s_{\min}(A)\leq s_{\max}(A)\leq \sqrt{N}+\sqrt{n}+t$$
\end{lemma}

\subsection{Perturbation Bound for Matrix Pseudo-inverse}
With a lowerbound on $\sigma_{\min}(A)$, we can get bounds for the perturbation of pseudo-inverse.
\begin{lemma}[Theorem 3.4 in~\cite{stewart1977perturbation}]
Consider the perturbation of a matrix $A\in \R^{m\times n}: B=A+E.$ Assume that $rank(A)=rank(B)=n,$ then
$$\n{B^\dagger - A^\dagger}\leq \sqrt{2}\n{A^\dagger}\n{B^\dagger}\n{E}.$$
\end{lemma}
The following corollary is particularly useful for us.
\begin{lemma}[Lemma G.8 in~\cite{ge2015learning}]\label{lem:inverse_perturb}
Consider the perturbation of a matrix $A\in \R^{m\times n}: B=A+E$ where $\n{E}\leq \sigma_{\min}(A)/2.$ Assume that $rank(A)=rank(B)=n,$ then
$$\n{B^\dagger-A^\dagger}\leq 2\sqrt{2}\n{E}/\sigma_{\min}(A)^2.$$
\end{lemma}

\end{document}